\documentclass[a4paper,twoside]{article}
\usepackage{graphicx}
\usepackage{grffile}

\usepackage[]{natbib}

\newif\ifscitepress
\scitepressfalse

\usepackage{calc}
\usepackage{amssymb}
\usepackage{amstext}
\usepackage{amsmath}
\usepackage{amsthm}
\usepackage{multicol}
\usepackage{pslatex}
\usepackage{apalike}

\usepackage{balance}
\usepackage{gensymb} %
\usepackage{arydshln}
\usepackage{booktabs}
\usepackage{multirow}
\usepackage[hyphens]{url}
\usepackage[colorlinks=true,allcolors=black]{hyperref}
\usepackage[acronym]{glossaries}
\usepackage[caption=false,farskip=0pt]{subfig}
\usepackage[table,dvipsnames]{xcolor}
\usepackage{xspace} %
\usepackage{diagbox}
\usepackage{soul} %
\setuldepth{Berlin}

\usepackage[utf8]{inputenc}
\usepackage[T1]{fontenc}

\newlength{\adj}
\newlength{\adjj}
\newcommand\wrong[1]{{\textcolor{red}{#1}}}

\newacronymstyle{long-short-br}
{%
  \GlsUseAcrEntryDispStyle{long-short}%
}%
{%
  \GlsUseAcrStyleDefs{long-short}%
}
\setacronymstyle{long-short-br}

\newcommand\blfootnote[1]{%
  \begingroup
  \renewcommand\thefootnote{}\footnote{#1}%
  \addtocounter{footnote}{-1}%
  \endgroup
}

\usepackage{SCITEPRESS}     %

\begin{document}

\title{On the Cross-Dataset Generalization in License Plate Recognition} 

\ifscitepress
\author{\authorname{Rayson Laroca\sup{1}\customorcidAuthor{0000-0003-1943-2711}, Everton V. Cardoso\sup{1}\customorcidAuthor{0000-0003-4845-6050}, Diego R. Lucio\sup{1}\customorcidAuthor{0000-0003-2012-3676},\\Valter Estevam\sup{1,2}\customorcidAuthor{0000-0001-9491-5882}, and David Menotti\sup{1}\customorcidAuthor{0000-0003-2430-2030}}
\affiliation{\sup{1}Federal University of Paran\'{a}, Curitiba, Brazil}
\affiliation{\sup{2}Federal Institute of Paran\'{a}, Irati, Brazil}
\email{\{rblsantos, evcardoso, drlucio, vlejunior, menotti\}@inf.ufpr.br, valter.junior@ifpr.edu.br}
}
\else
\author{\authorname{Rayson Laroca\sup{1}\customorcidAuthorA{0000-0003-1943-2711}, Everton V. Cardoso\sup{1}\customorcidAuthorA{0000-0003-4845-6050}, Diego R. Lucio\sup{1}\customorcidAuthorA{0000-0003-2012-3676},\\Valter Estevam\sup{1,2}\customorcidAuthorA{0000-0001-9491-5882}, and David Menotti\sup{1}\customorcidAuthorA{0000-0003-2430-2030}}
\affiliation{\sup{1}Federal University of Paran\'{a}, Curitiba, Brazil}
\affiliation{\sup{2}Federal Institute of Paran\'{a}, Irati, Brazil}
\email{\{rblsantos, evcardoso, drlucio, vlejunior, menotti\}@inf.ufpr.br, valter.junior@ifpr.edu.br}
}

\fi


\newacronym{alpr}{ALPR}{Automatic License Plate Recognition}
\newacronym{bilstm}{Bi-LSTM}{Bi-directional Long Short-Term Memory}
\newacronym{cnn}{CNN}{Convolutional Neural Network}
\newacronym{denatran}{DENATRAN}{National Traffic Department of Brazil}
\newacronym{gan}{GAN}{Generative Adversarial Network}
\newacronym{lp}{LP}{license plate}
\newacronym{lodo}{LODO}{leave-one-dataset-out}
\newacronym{lstm}{LSTM}{Long Short-Term Memory}
\newacronym{iou}{IoU}{Intersection over Union}
\newacronym{it}{IT}{information technology}
\newacronym{ocr}{OCR}{Optical Character Recognition}
\newacronym{rodosol}{RodoSol}{\textit{Rodovia do Sol}}
\newacronym{sgd}{SGD}{Stochastic Gradient Descent}
\newacronym{spp}{SPP}{Spatial Pyramid Pooling}

\newcommand{\dataset}{RodoSol-ALPR\xspace}
\newcommand{\xception}{Xception\xspace}

\newcommand{\aolp}{AOLP\xspace}
\newcommand{\caltech}{Caltech Cars\xspace}
\newcommand{\ccpd}{CCPD\xspace}
\newcommand{\chineselp}{ChineseLP\xspace}
\newcommand{\clpd}{CLPD\xspace}
\newcommand{\englishlp}{EnglishLP\xspace}
\newcommand{\karplate}{KarPlate\xspace}
\newcommand{\openalpreu}{OpenALPR-EU\xspace}
\newcommand{\platesmaniaiet}{PlatesMania\xspace}
\newcommand{\ssigsegplate}{SSIG-SegPlate\xspace}
\newcommand{\stills}{UCSD-Stills\xspace}
\newcommand{\ufpralpr}{UFPR-ALPR\xspace}

\newcommand{\crnn}{CRNN\xspace}
\newcommand{\grcnn}{GRCNN\xspace}
\newcommand{\rare}{RARE\xspace}
\newcommand{\rosetta}{Rosetta\xspace}
\newcommand{\rtwoam}{R\textsuperscript{2}AM\xspace}
\newcommand{\starnet}{STAR-Net\xspace}
\newcommand{\trba}{TRBA\xspace}
\newcommand{\vitstrbase}{ViTSTR-Base\xspace}

\newcommand{\holistic}{Holistic-CNN\xspace}
\newcommand{\multitaskgabriel}{Multi-task\xspace}

\newcommand{\crnet}{CR-NET\xspace}
\newcommand{\fastocr}{Fast-OCR\xspace}

\newcommand{\numimages}{20{,}000\xspace}
\newcommand{\numkm}{67.5\xspace}

\newcommand{\numdatasets}{9\xspace}
\newcommand{\numbaselines}{12\xspace}
\newcommand{\numcommercial}{2\xspace}
\newcommand{\nummodelstrained}{120\xspace}

\newcommand{\acclodo}{74.5\xspace}
\newcommand{\acctraditional}{82.4\xspace}

\newcommand{\supplementary}{\url{https://github.com/raysonlaroca/rodosol-alpr-dataset/}}
\newcommand{\discarded}{\url{https://raysonlaroca.github.io/supp/visapp2022/discarded-images.txt}}
\abstract{
\gls*{alpr} systems have shown remarkable performance on \glspl*{lp} from multiple regions due to advances in deep learning and the increasing availability of datasets.
The evaluation of deep \gls*{alpr} systems is usually done within each dataset; therefore, it is questionable if such results are a reliable indicator of generalization ability.
In this paper, we propose a traditional-split \textit{versus} leave-one-dataset-out experimental setup to empirically assess the cross-dataset generalization of $12$ \gls*{ocr} models applied to \gls*{lp} recognition on nine publicly available datasets with a great variety in several aspects (e.g., acquisition settings, image resolution, and \gls*{lp} layouts).
We also introduce a public dataset for end-to-end \gls*{alpr} that is the first to contain images of vehicles with Mercosur \glspl*{lp} and the one with the highest number of motorcycle images.
The experimental results shed light on the limitations of the traditional-split protocol for evaluating approaches in the \gls*{alpr} context, as there are significant drops in performance for most datasets when training and testing the models in a leave-one-dataset-out~fashion.
\ifscitepress
\else
\vspace{3mm}
\fi
}

\onecolumn \maketitle \normalsize \setcounter{footnote}{0} \vfill

\section{\uppercase{Introduction}}
\label{sec:introduction}

\glsresetall

The global automotive industry expects to produce more than $82$ million light vehicles in 2022 alone, despite the ongoing coronavirus pandemic and chip supply issues~\citep{news2021forbes,news2021ihsmarkit}.
In addition to bringing convenience to owners, vehicles also significantly modify the urban environment, posing challenges concerning pollution, privacy and security --~especially in large urban centers.
The constant monitoring of vehicles through computational techniques is of paramount importance and, therefore, it has been a frequent research topic.
In this context, \gls*{alpr} systems~\citep{weihong2020research,lubna2021automatic} stand out.%
\ifscitepress
\else
\blfootnote{\scriptsize This is an author-prepared version of a paper accepted for presentation at the International Conference on Computer Vision Theory and Applications (VISAPP) 2022. The published version is available at the \emph{SciTePress Digital Library} (DOI: \href{https://doi.org/10.5220/0010846800003124}{\textcolor{blue}{10.5220/0010846800003124}}).}
\fi

\gls*{alpr} systems exploit image processing and pattern recognition techniques to detect and recognize the characters on \glspl*{lp} from images or videos.
Some practical applications for an \gls*{alpr} system are road traffic monitoring, toll collection, and vehicle access control in restricted areas~\citep{spanhel2017holistic,henry2020multinational,wang2022rethinking}.

Deep \gls*{alpr} systems have shown remarkable performance on \glspl*{lp} from multiple regions due to advances in deep learning and the increasing availability of datasets~\citep{henry2020multinational,silva2022flexible}.
In the past, the evaluation of \gls*{alpr} systems used to be done within each of the chosen datasets, i.e., the proposed methods were trained and evaluated on different subsets from the same dataset.
Such an evaluation was carried out independently for each dataset.
Recently, considering that deep models can take considerable time to be trained (especially on low- or mid-end GPUs), the authors have adopted a protocol where the proposed models are trained once on the union of the training images from the chosen datasets and evaluated individually on the respective test sets~\citep{selmi2020delpdar,laroca2021efficient}.
Although the images for training and testing belong to disjoint subsets, these protocols do not make it clear whether the evaluated models have good generalization ability, i.e., whether they perform well on images from other scenarios, mainly due to domain divergence and data selection bias~\citep{torralba2011unbiased,tommasi2017deeper,zhang2019recent}.

In this regard, many computer vision researchers have carried out cross-dataset experiments --~where training and testing data come from different sources~-- to assess whether the proposed models perform well on data from an unknown domain~\citep{ashraf2018learning,zhang2019recent,estevam2021tell}.
However, as far as we know, there is no work focused on such experimental settings in the \gls*{alpr} context.

Considering the above discussion, in this work we evaluate for the first time various \gls*{ocr} models for \gls*{lp} recognition in a leave-one-dataset-out experimental setup over nine public datasets with different characteristics.
The results obtained are compared with those achieved when training the models in the same way as in recent works, that is, using the union of the training set images from all datasets (hereinafter, this protocol is referred to as traditional-split). 

Deep learning-based \gls*{alpr} systems have often achieved recognition rates above $99$\% in existing datasets under the traditional-split protocol (some examples are provided in Section~\ref{sec:related_work}). However, in real-world applications, new cameras are regularly being installed in new locations without existing systems being retrained as often, which can dramatically decrease the performance of those models.
A leave-one-dataset-out protocol enables simulating this specific scenario and providing an adequate evaluation of the generalizability of the models.

\gls*{alpr} is commonly divided into two tasks: \gls*{lp} detection and \gls*{lp} recognition.
The former refers to locating the \gls*{lp} region in the input image, while the latter refers to extracting the string related to the \gls*{lp}.
In this work, we focus on the \gls*{lp} recognition stage since it is the current bottleneck of \gls*{alpr} systems~\citep{laroca2021efficient}.
Thus, we simply train the off-the-shelf YOLOv4 model~\citep{bochkovskiy2020yolov4} to detect the \glspl*{lp} in the input images.
For completeness, we also report the results achieved in this stage on both of the aforementioned~protocols.

As part of this work, we introduce a publicly available dataset, called \dataset\footnote{The \dataset dataset is publicly available to the research community at \supplementary}, that contains $\numimages$ images captured at toll booths installed on a Brazilian highway.
It has images of two different \gls*{lp} layouts: Brazilian and Mercosur\footnote{Mercosur (\textit{Mercado Com\'{u}n del Sur}, i.e., Southern Common Market in Castilian) is an economic and political bloc
comprising Argentina, Brazil, Paraguay and Uruguay.}, with half of the vehicles being motorcycles (see details in Section~\ref{sec:dataset}).
To the best of our knowledge, this is the first public dataset for \gls*{alpr} with images of Mercosur \glspl*{lp} and the largest in the number of motorcycle images.
This last information is relevant because motorcycle \glspl*{lp} have two rows of characters, which is a challenge for sequential/recurrent-based methods~\citep{silva2022flexible}, and therefore have been overlooked in the evaluation of \gls*{lp} recognition models (see Section~\ref{sec:related_work}).

Our paper has two main contributions:
\begin{itemize}
    \item A traditional-split \textit{versus} leave-one-dataset-out experimental setup that can be considered a valid testbed for cross-dataset generalization methods proposed in future works on \gls*{alpr}.
    We present a comparative assessment of $\numbaselines$ \gls*{ocr} models for \gls*{lp} recognition on nine publicly available datasets.
    The main findings were that
    (i)~there are significant drops in performance for most datasets when training and testing the recognition models in a leave-one-dataset-out fashion, especially when there are different fonts of characters in the training and test images;
    (ii)~no model achieved the best result in all experiments, with $6$ different models reaching the best result in at least one dataset under the leave-one-dataset-out protocol; 
    and (iii)~the proposed dataset proved very challenging, as both the models trained by us and two commercial systems failed to reach recognition rates above $70$\% on its test set~images.
    \item A public dataset with $\numimages$ images acquired in real-world scenarios, being half of them of vehicles with Mercosur \glspl*{lp}.
    Indeed, one of the objectives of this work is to provide a reliable source of information about Mercosur \glspl*{lp}, as much news --~often outdated~-- has been used as~references.
\end{itemize}

The remainder of this paper is organized in the following manner. 
In Section~\ref{sec:related_work}, we briefly review related works.
The \dataset dataset is introduced in Section~\ref{sec:dataset}.
The setup adopted in our experiments is thoroughly described in Section~\ref{sec:experiments}.
Section~\ref{sec:results} presents the results achieved.
Finally, Section~\ref{sec:conclusions} concludes the paper and outlines future directions of~research.
\section{\uppercase{Related Work}}
\label{sec:related_work}

In this section, we first present  concisely recent works on \gls*{lp} recognition.
Then, we situate the current state of \gls*{alpr} research in terms of cross-dataset experiments, Mercosur \glspl*{lp}, and motorcycle~\glspl*{lp}.

The good speed/accuracy trade-off provided by YOLO networks~\citep{redmon2016yolo,bochkovskiy2020yolov4} inspired many authors to explore similar architectures targeting real-time performance for \gls*{lp} recognition.
For example, \cite{silva2020realtime} proposed a YOLO-based model to simultaneously detect and recognize all characters within a cropped \gls*{lp}. This model, called \crnet, consists of the first eleven layers of YOLO and four other convolutional layers added to improve non-linearity.
Impressive results were achieved through \crnet both in the original work and in more recent ones~\citep{laroca2021efficient,oliveira2021vehicle,silva2022flexible}.

While \cite{kessentini2019twostage} applied the YOLOv2 model without any change or refinement to this task, \cite{henry2020multinational} used a modified version of YOLOv3 that includes spatial pyramid pooling.
Although these two models achieved high recognition rates in multiple datasets, they are very deep for \glspl*{lp} recognition, making it difficult to meet the real-time requirements of \gls*{alpr}~applications.

Rather than exploring object detectors, \cite{zou2020robust} adopted a bi-directional \gls*{lstm} network to implicitly locate the characters on the \gls*{lp}.
They explored a $1$-D attention module to extract useful features of the character regions, improving the accuracy of \gls*{lp} recognition.
In a similar way, \cite{zhang2021robust_attentional} used a $2$-D attention mechanism to optimize their recognition model, which uses a 30-layer \gls*{cnn} based on \xception for feature~extraction.
An \gls*{lstm} model was adopted to decode the extracted features into \gls*{lp} characters.

There are also several works where multi-task networks were designed to holistically process the entire \gls*{lp} image and, thus, avoid character segmentation, such as~\citep{spanhel2017holistic,goncalves2019multitask}.
As these networks employ fully connected layers as classifiers to recognize the characters on the predefined positions of the \glspl*{lp}, they may not generalize well with small-scale training sets since the probability of a specific character appearing in a specific position is low.
To deal with this, \cite{wang2022rethinking} proposed a weight-sharing classifier, which is able to spot instances of each character across all positions.

Considering that the recognition rates achieved under the traditional-split protocol have significantly increased in recent years, some authors began to conduct small cross-dataset experiments to analyze the generalization ability of the proposed methods.
For example, \cite{silva2020realtime,laroca2021efficient} used all $108$ images from the \openalpreu dataset for testing, rather than using some for training/validation.
Nevertheless, the results achieved in so few test images are susceptible to tricks, especially considering that heuristic rules were explored to improve the \gls*{lp} recognition results in both~works.

As another example, \cite{zou2020robust,zhang2021robust_attentional,wang2022rethinking} trained their recognition models specifically for Chinese \glspl*{lp} on approximately $200$K images from the \ccpd dataset~\citep{xu2018towards} and tested them on images from other datasets that also contain only Chinese \glspl*{lp}.
In this case, it is not clear whether the proposed models perform well on \glspl*{lp} from other regions.
In fact, the authors trained another instance of the respective models to evaluate them in the \aolp dataset~\citep{hsu2013application}, which contains \glspl*{lp} from the Taiwan~region.

Recently, Mercosur countries adopted a unified standard of \glspl*{lp} for newly purchased vehicles, inspired by the integrated system adopted by European Union countries many years ago.
Although the new standard has been implemented in all countries in the bloc, there is still no public dataset for \gls*{alpr} with images of Mercosur \glspl*{lp} as far as we~know.

In this sense, \cite{silvano2021synthetic} presented a methodology that couples synthetic images of Mercosur \glspl*{lp} with real-world images containing vehicles with other \gls*{lp} layouts.
A model trained exclusively with synthetic images achieved promising results on $1{,}000$ real images from various sources; however, it is difficult to assess these results accurately since the test images were not made available to the research~community. 
The \gls*{lp} recognition stage was not~addressed.

Despite the fact that motorcycles are one of the most popular transportation means in metropolitan areas~\citep{hsu2015comparison}, they have been largely overlooked in \gls*{alpr} research.
There are even works where images of motorcycles were excluded from the experiments~\citep{goncalves2018realtime,silva2020realtime}, mainly because \glspl*{lp} of motorcycles usually have two rows of characters, 
which are challenging to sequential/recurrent-based methods~\citep{kessentini2019twostage,silva2022flexible}, and also because they are generally smaller in size (having less space between characters) and are often~tilted.

In this regard, there is a great demand for a public dataset for end-to-end \gls*{alpr} with the same number of images of cars and motorcycles to give equal importance to \glspl*{lp} with one or two rows of characters in the assessment of \gls*{alpr}~systems.
\begin{figure*}[!htb]
    \centering
        
    \resizebox{0.925\linewidth}{!}{
        \includegraphics[width=0.19\linewidth]{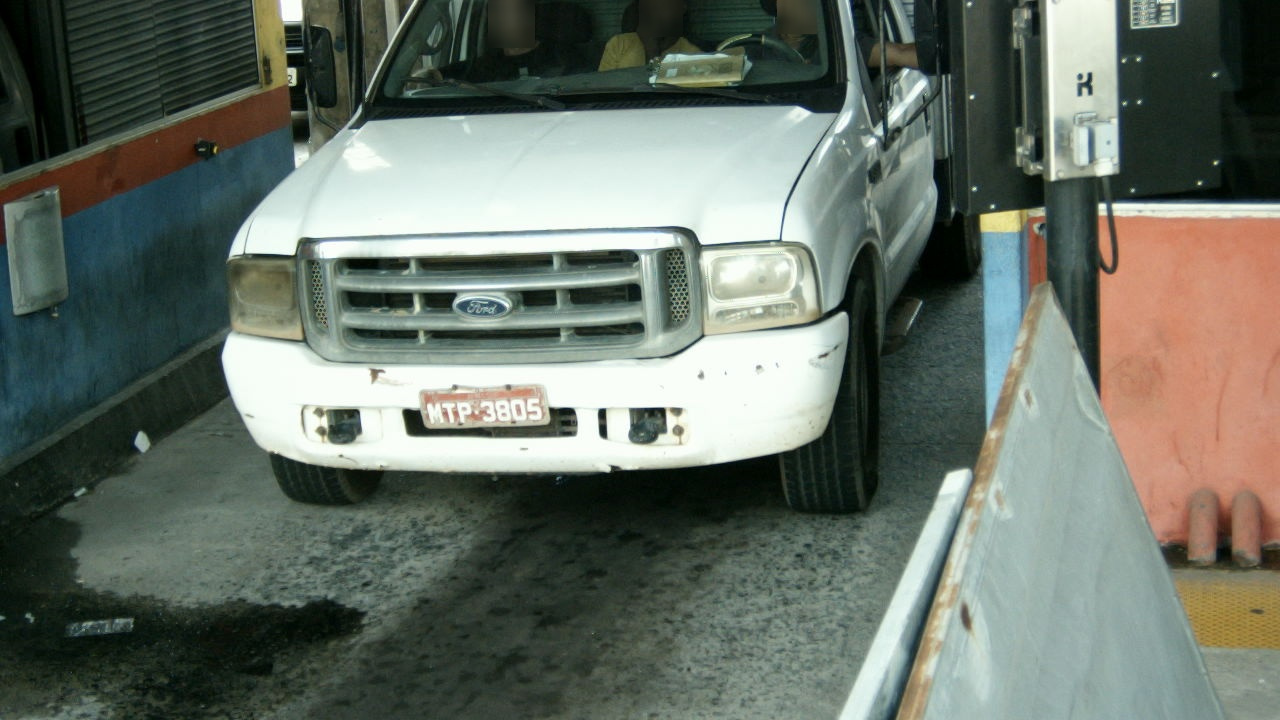}
        \includegraphics[width=0.19\linewidth]{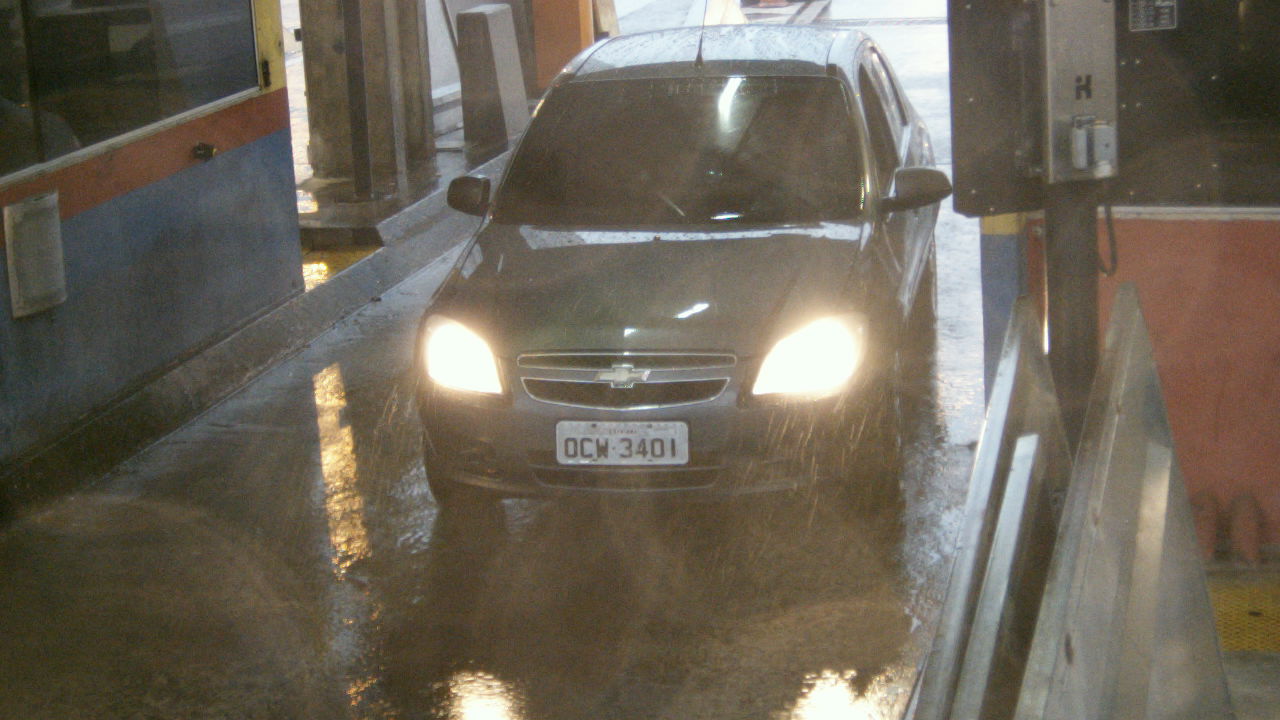}
        \includegraphics[width=0.19\linewidth]{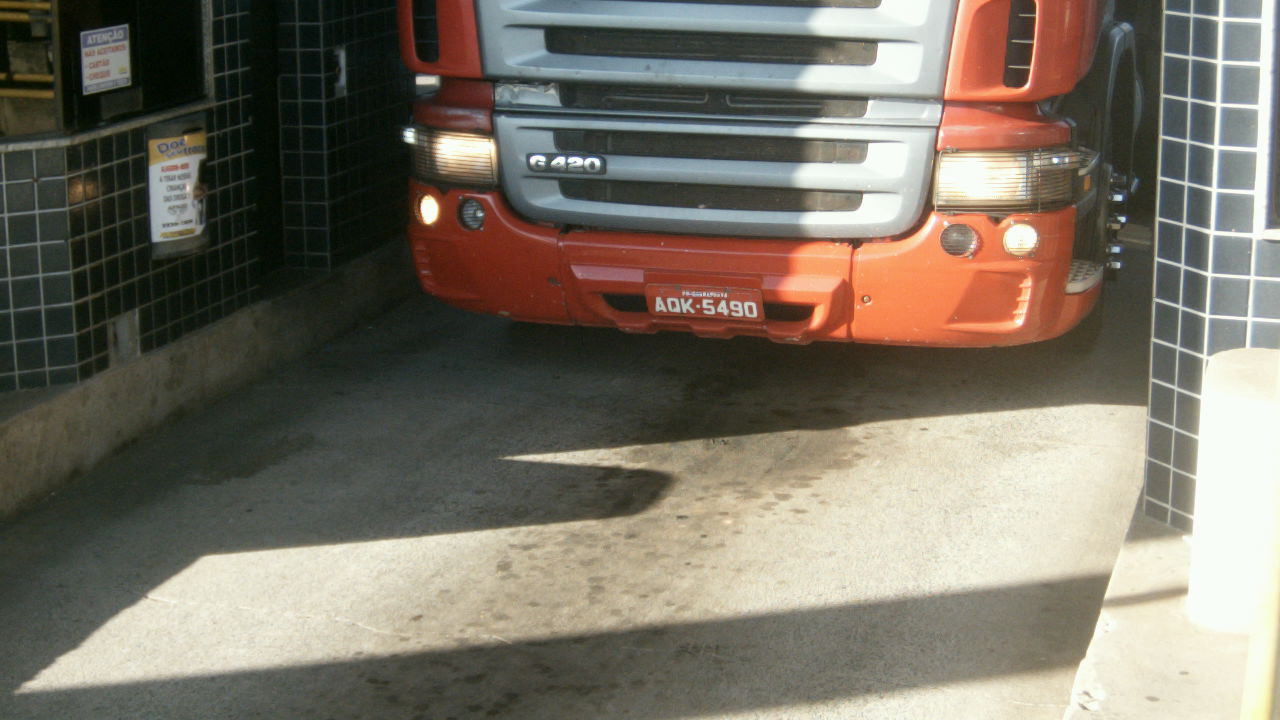}
        \includegraphics[width=0.19\linewidth]{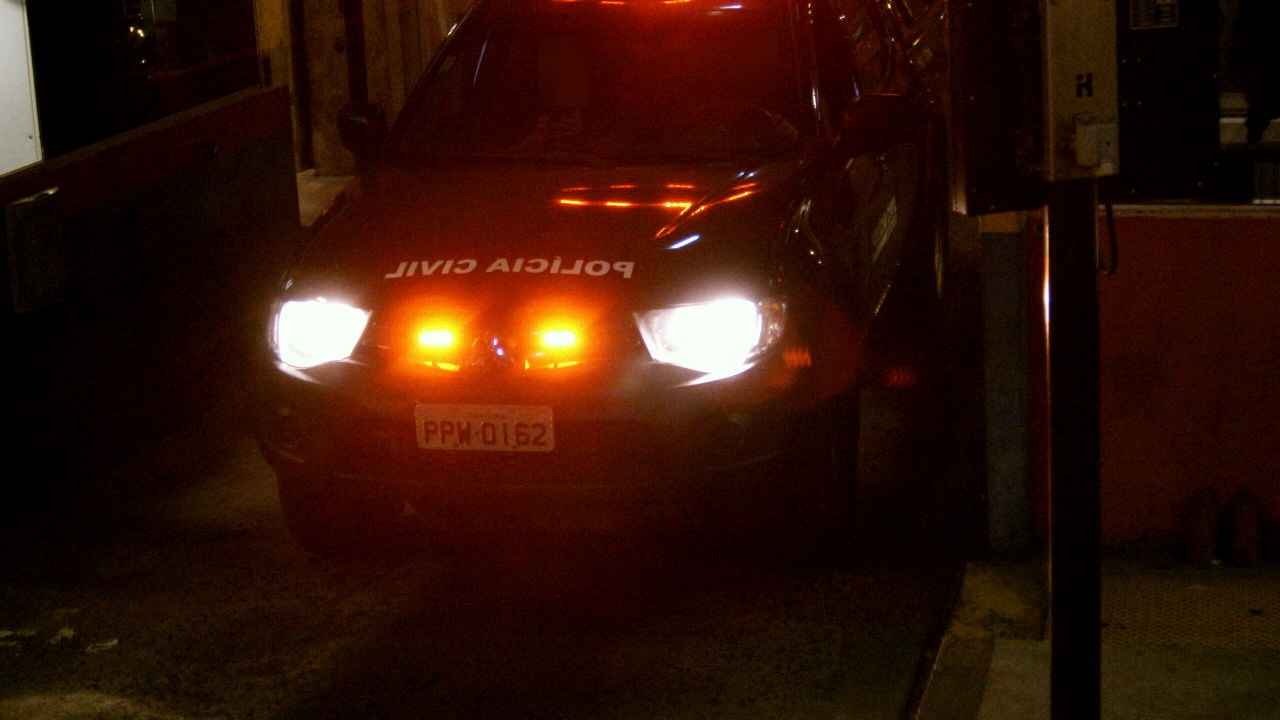}
        \includegraphics[width=0.19\linewidth]{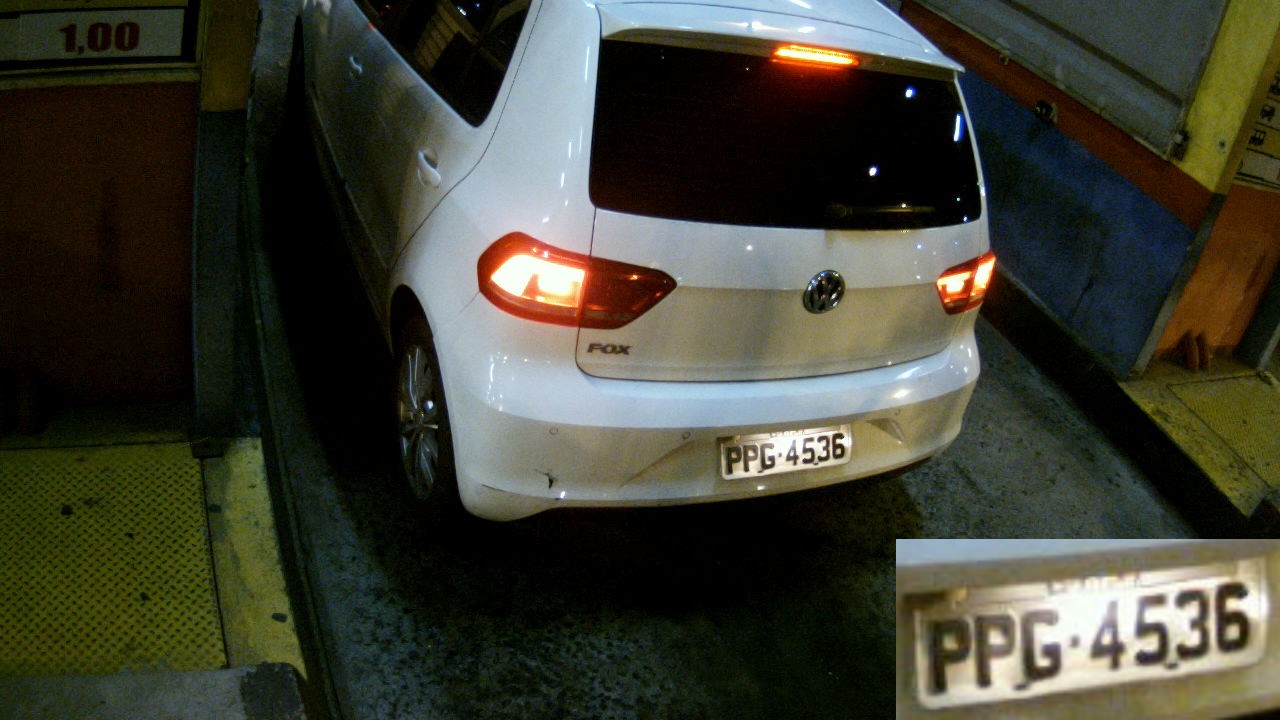}
    }
    
    \vspace{0.32mm}
    
    \resizebox{0.925\linewidth}{!}{
        \includegraphics[width=0.19\linewidth]{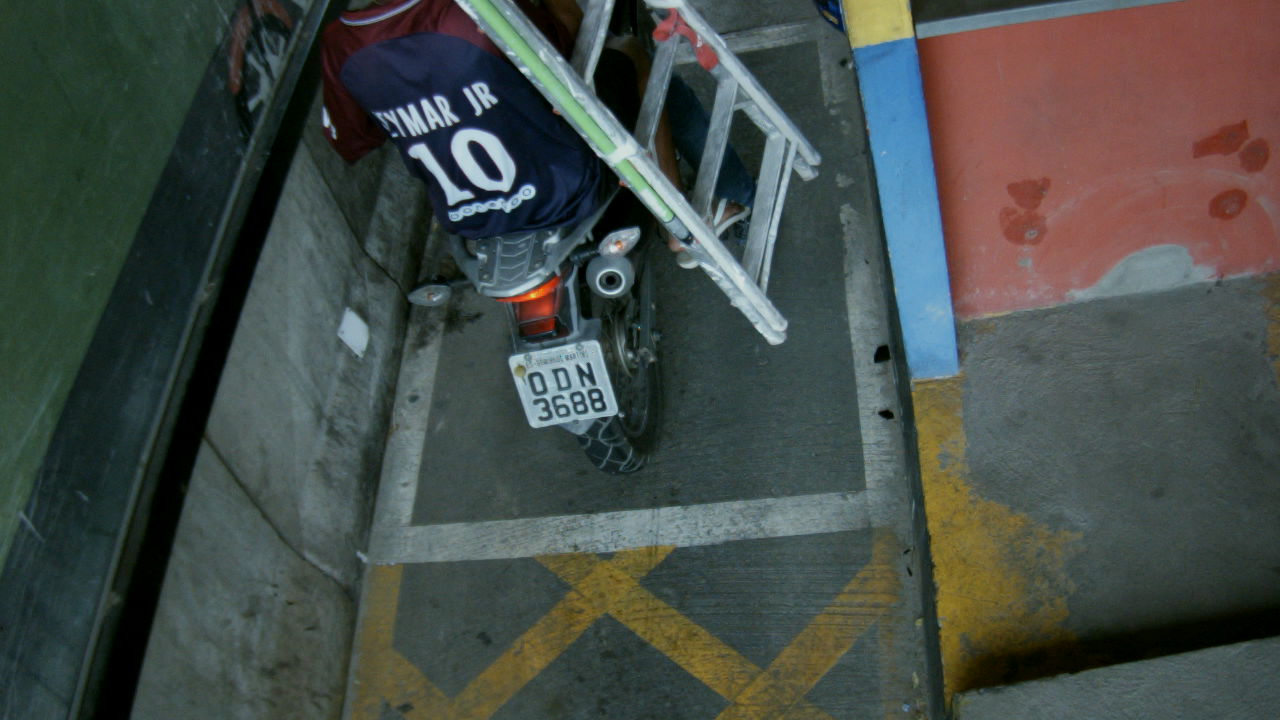}
        \includegraphics[width=0.19\linewidth]{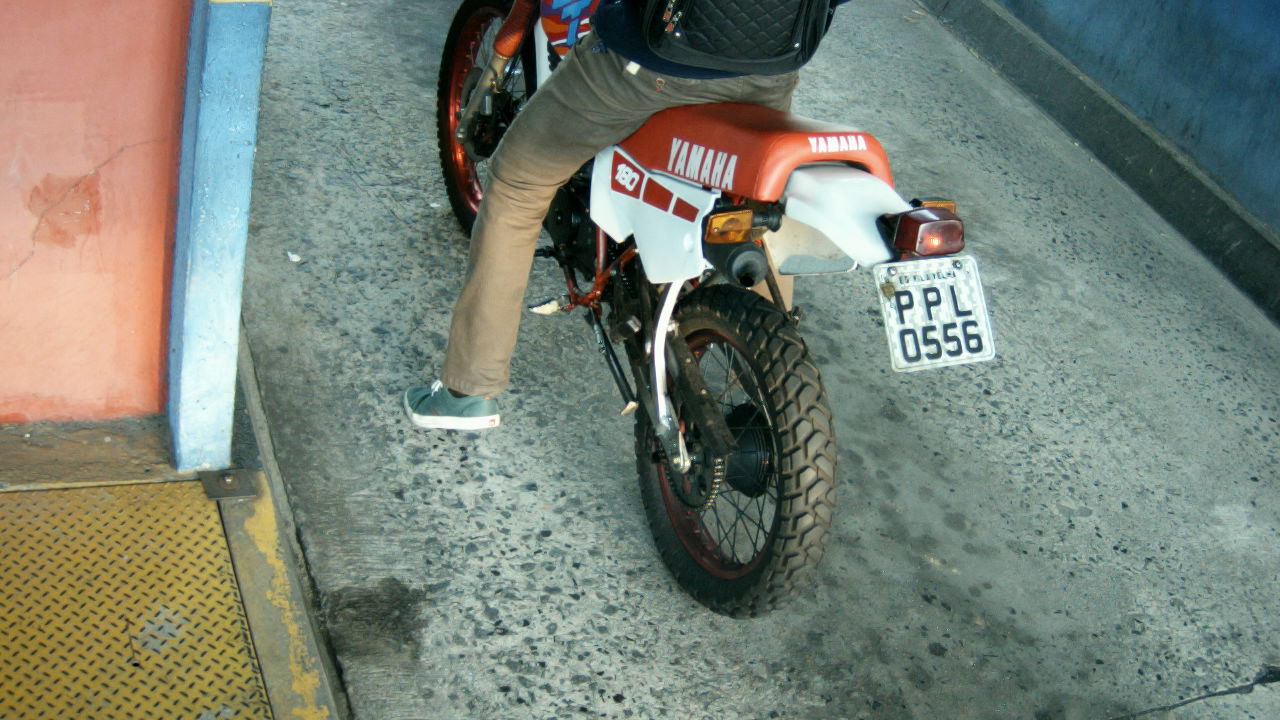}
        \includegraphics[width=0.19\linewidth]{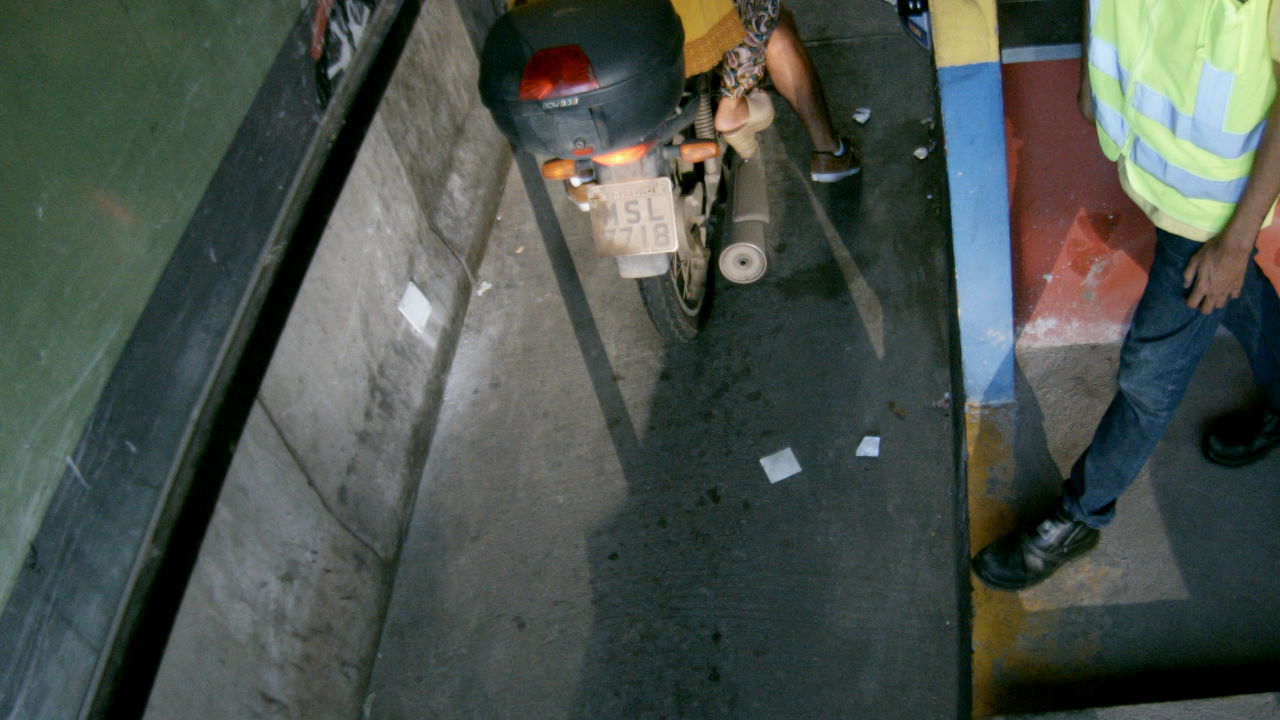}
        \includegraphics[width=0.19\linewidth]{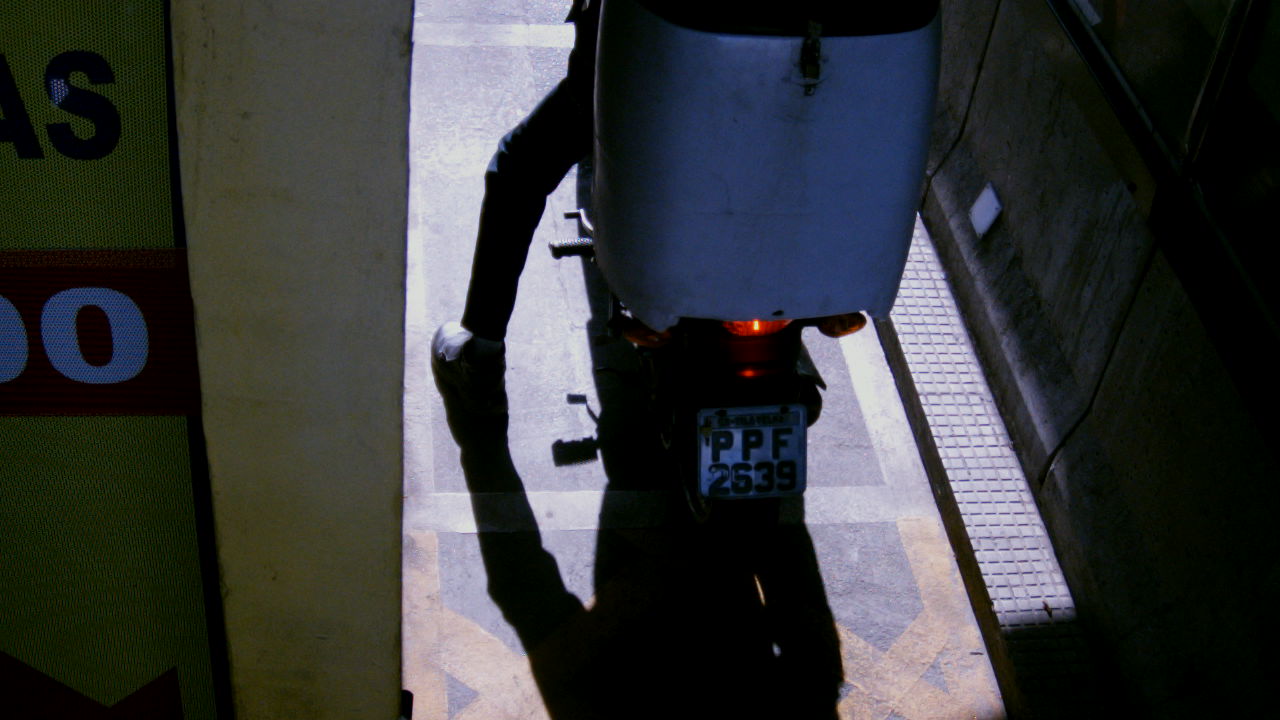}
        \includegraphics[width=0.19\linewidth]{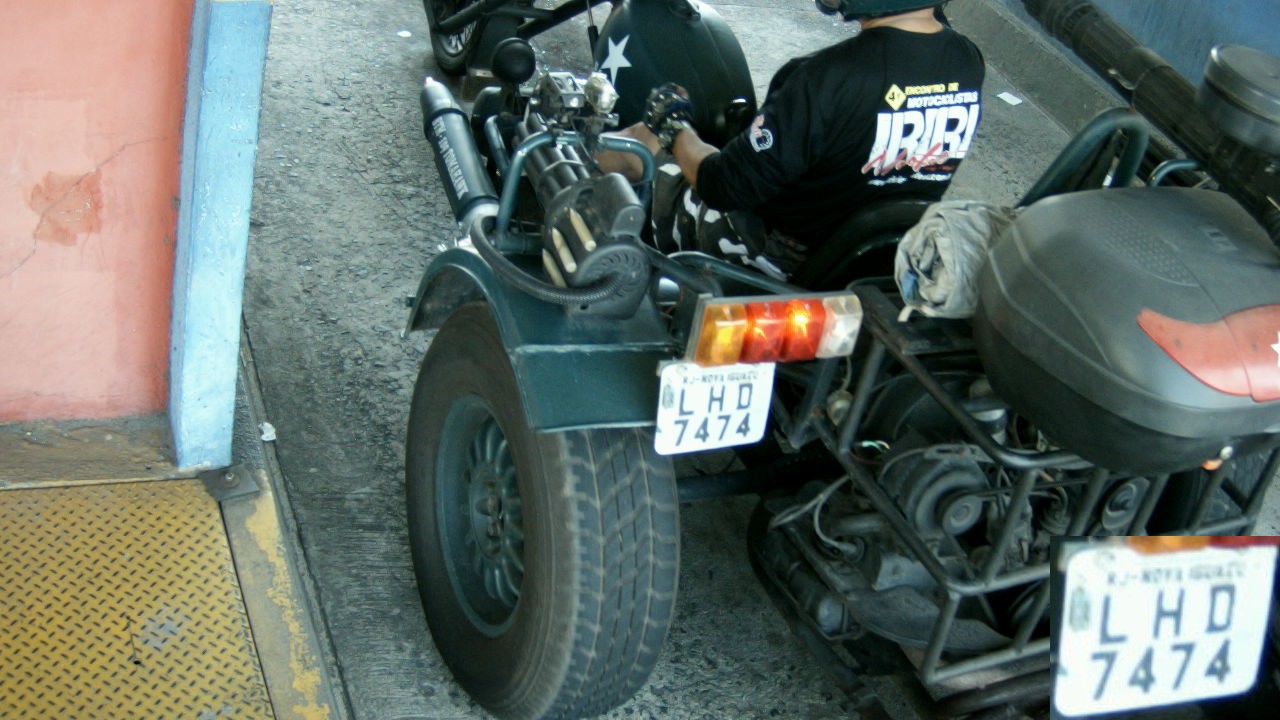}
    }
    
    \vspace{0.3mm}
    
    \resizebox{0.925\linewidth}{!}{
        \includegraphics[width=0.19\linewidth]{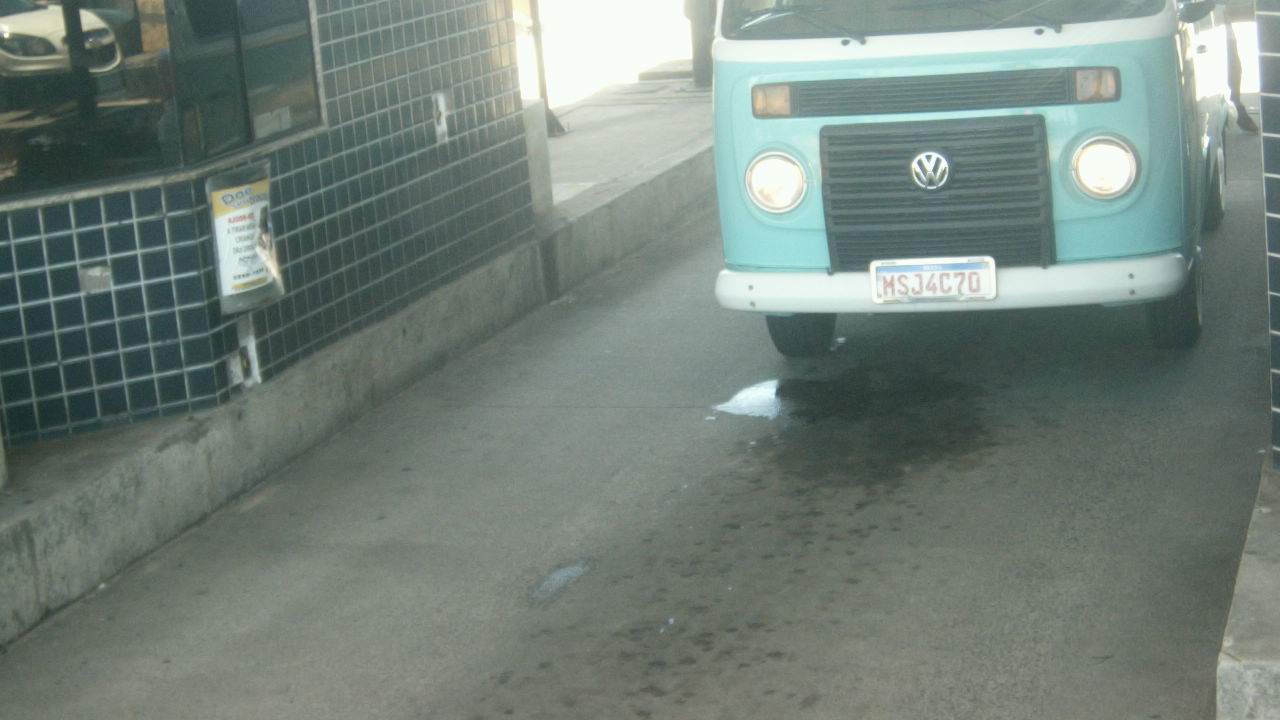}
        \includegraphics[width=0.19\linewidth]{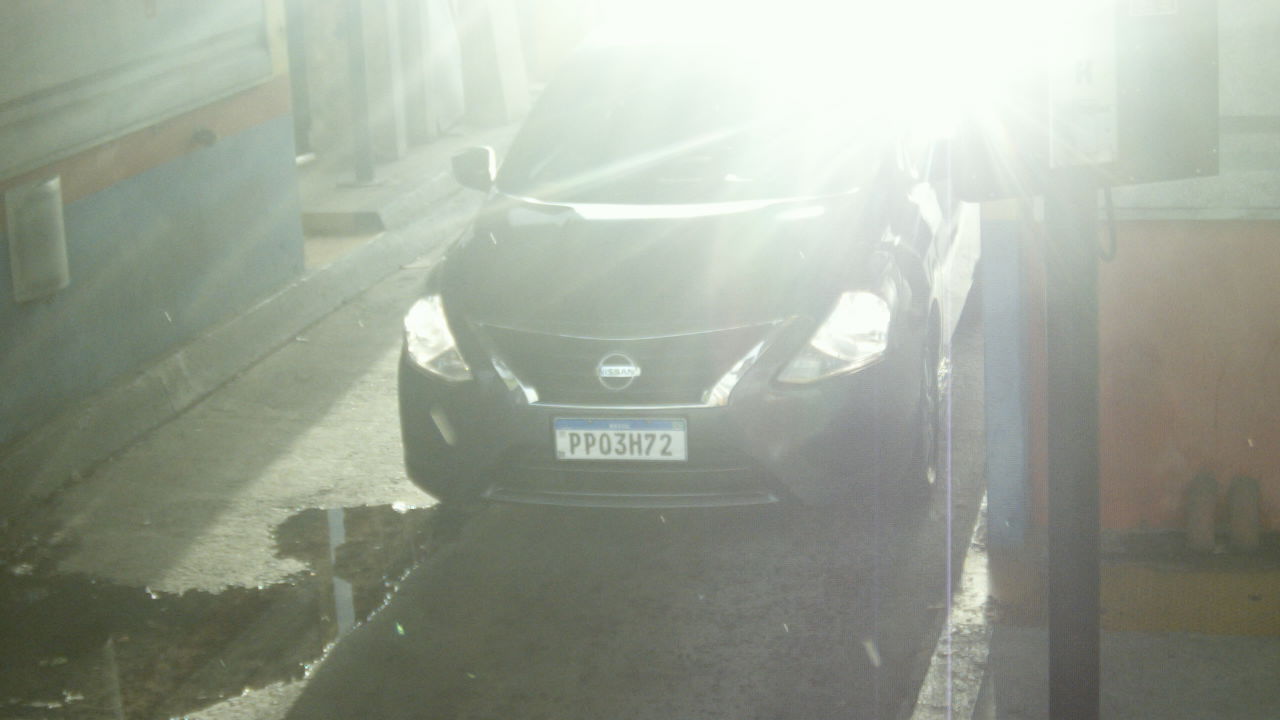}
        \includegraphics[width=0.19\linewidth]{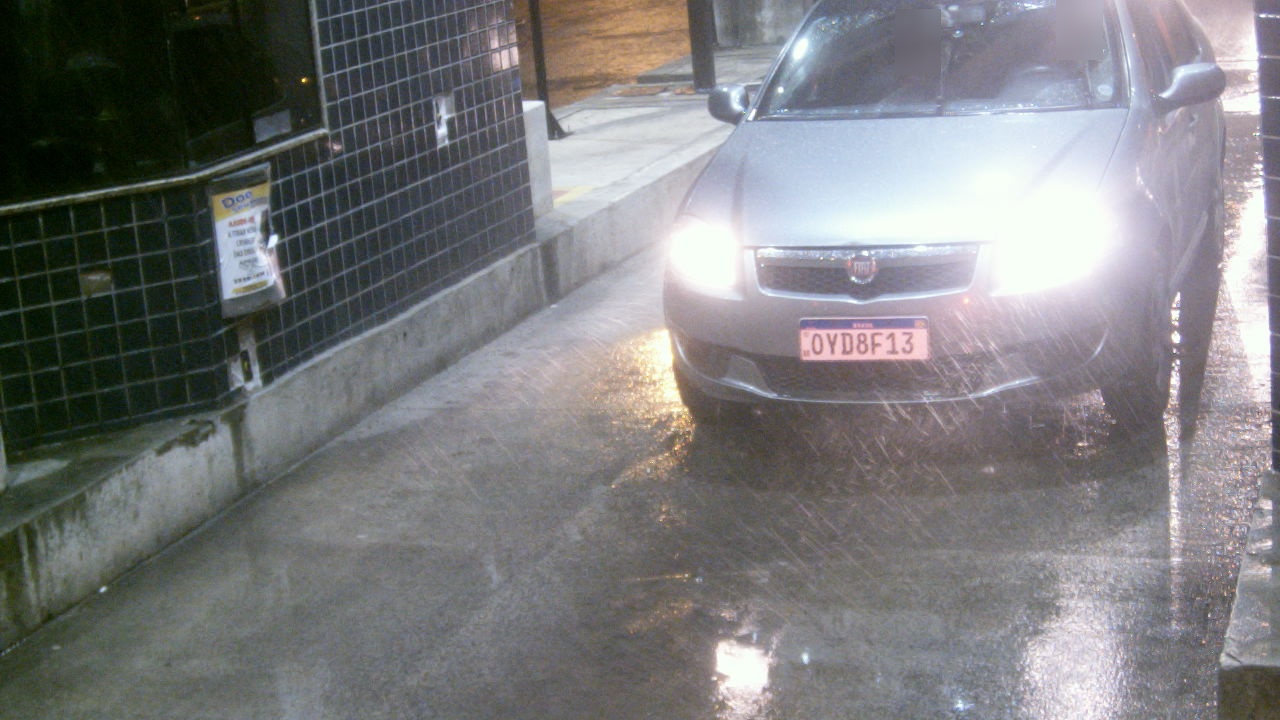}
        \includegraphics[width=0.19\linewidth]{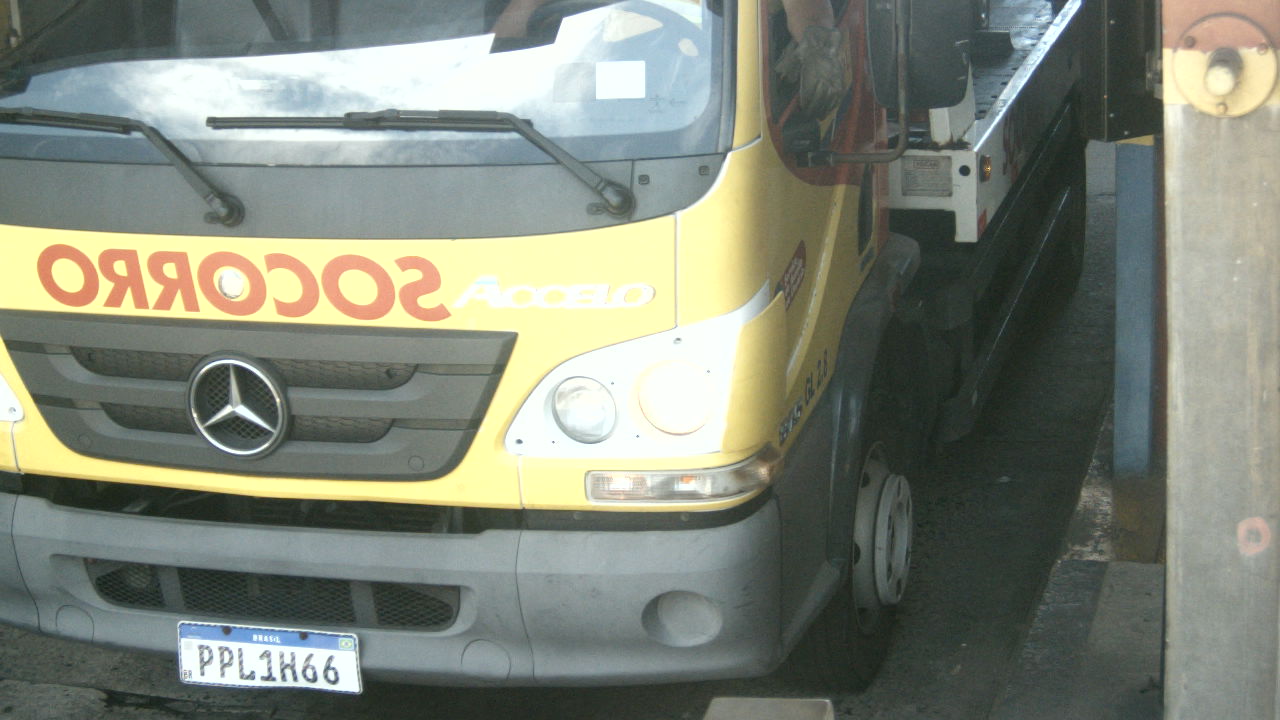}
        \includegraphics[width=0.19\linewidth]{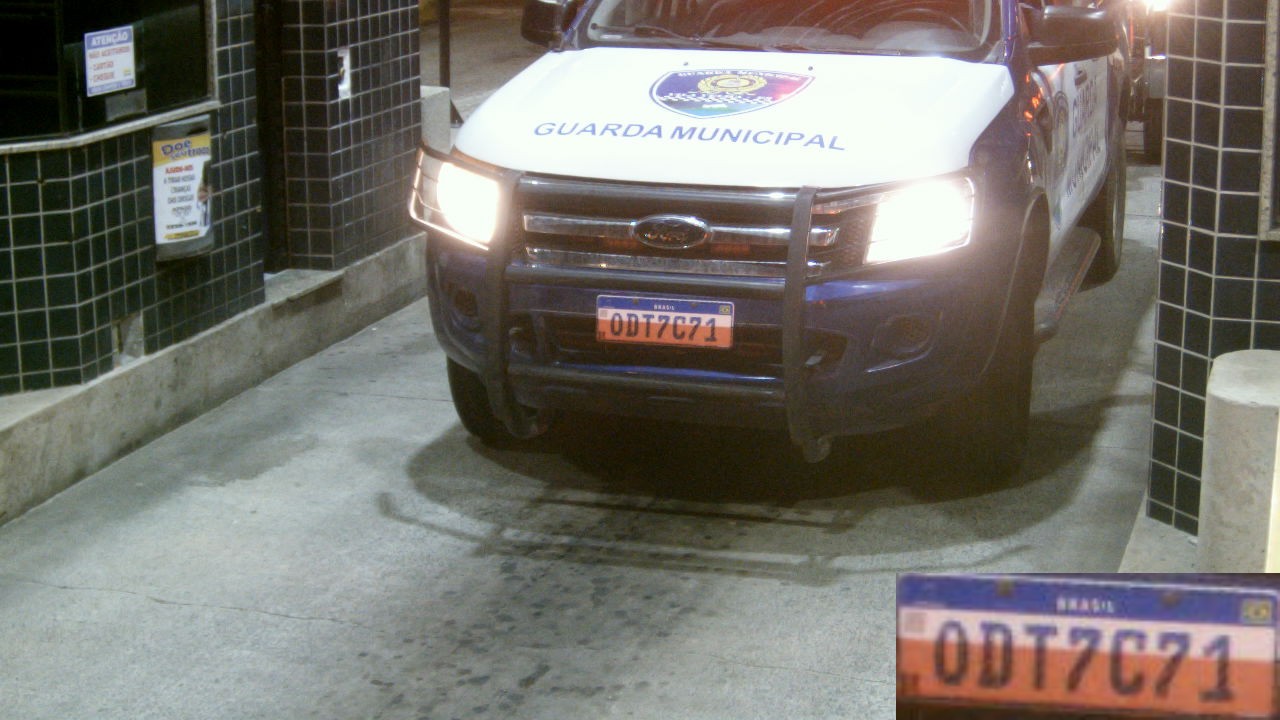}
    }
    
    \vspace{0.31mm}
    
    \resizebox{0.925\linewidth}{!}{
        \includegraphics[width=0.19\linewidth]{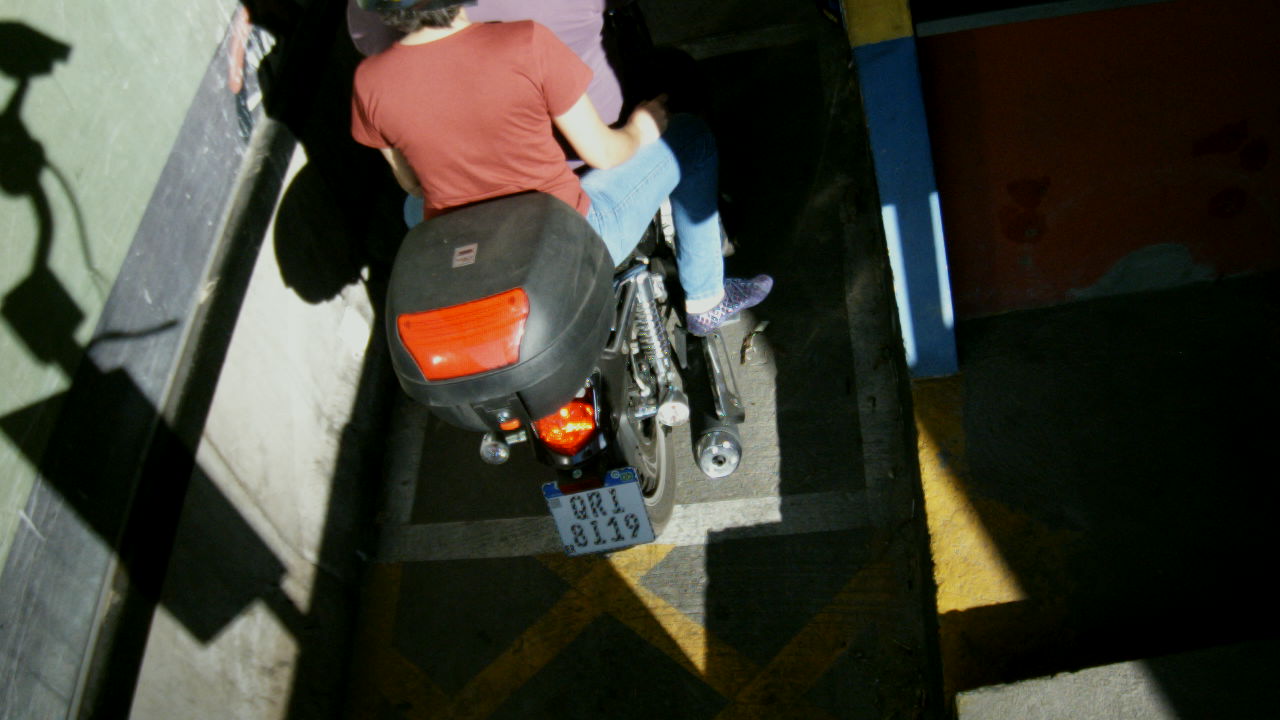}
        \includegraphics[width=0.19\linewidth]{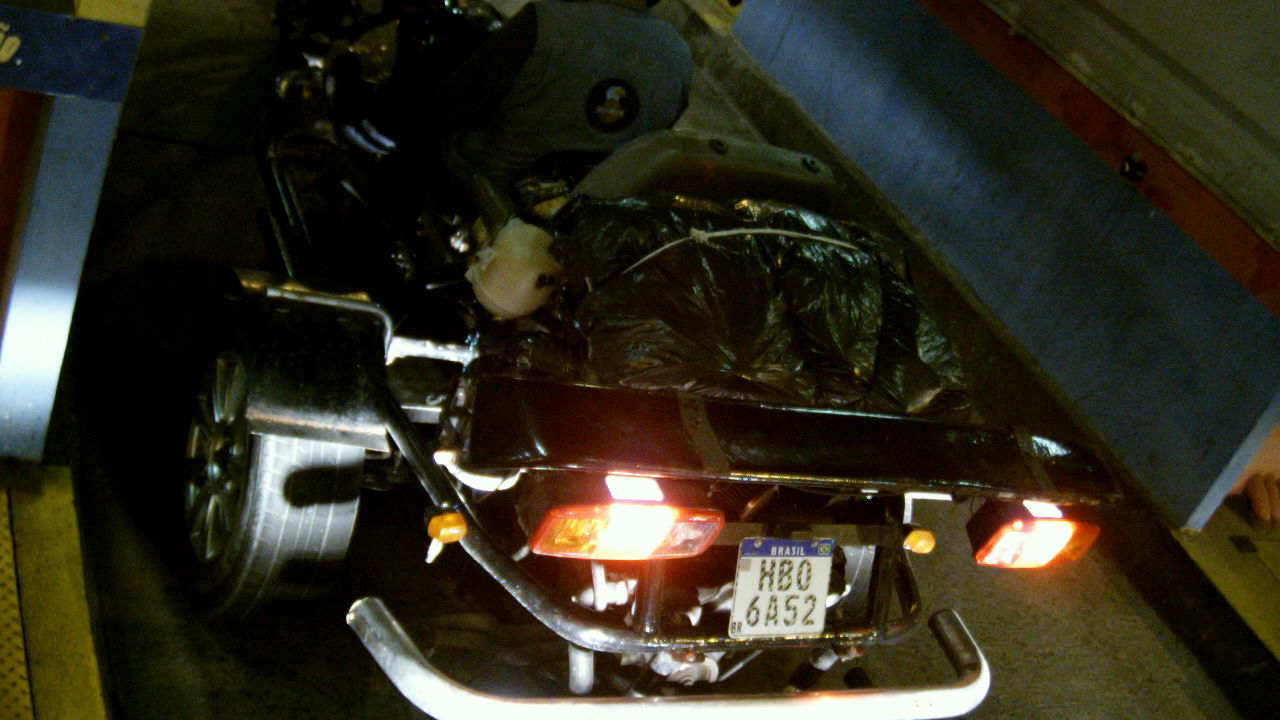}
        \includegraphics[width=0.19\linewidth]{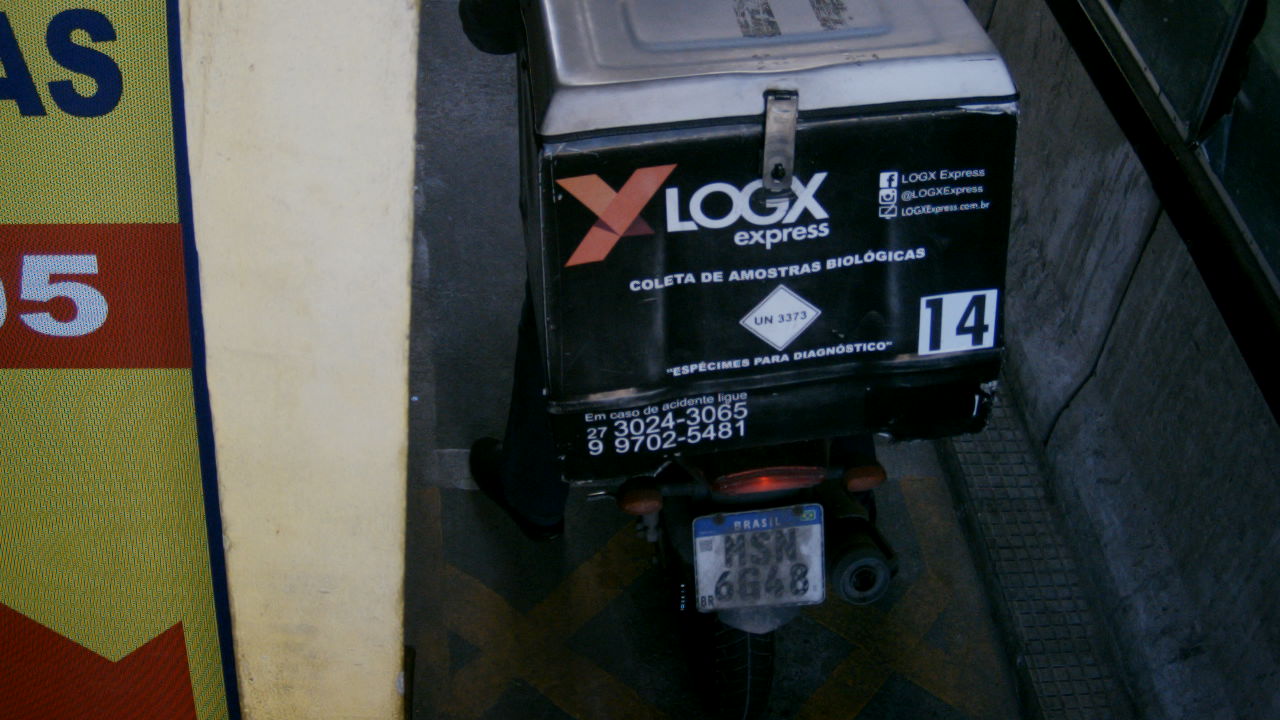}
        \includegraphics[width=0.19\linewidth]{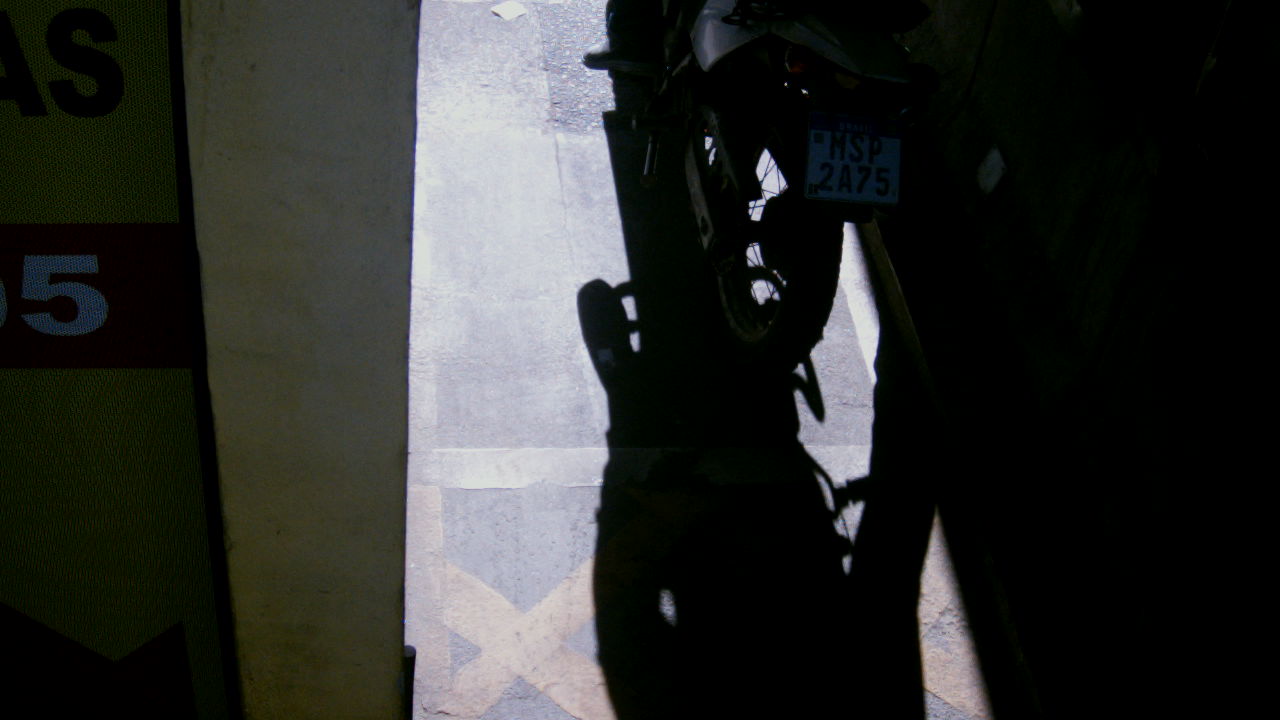}
        \includegraphics[width=0.19\linewidth]{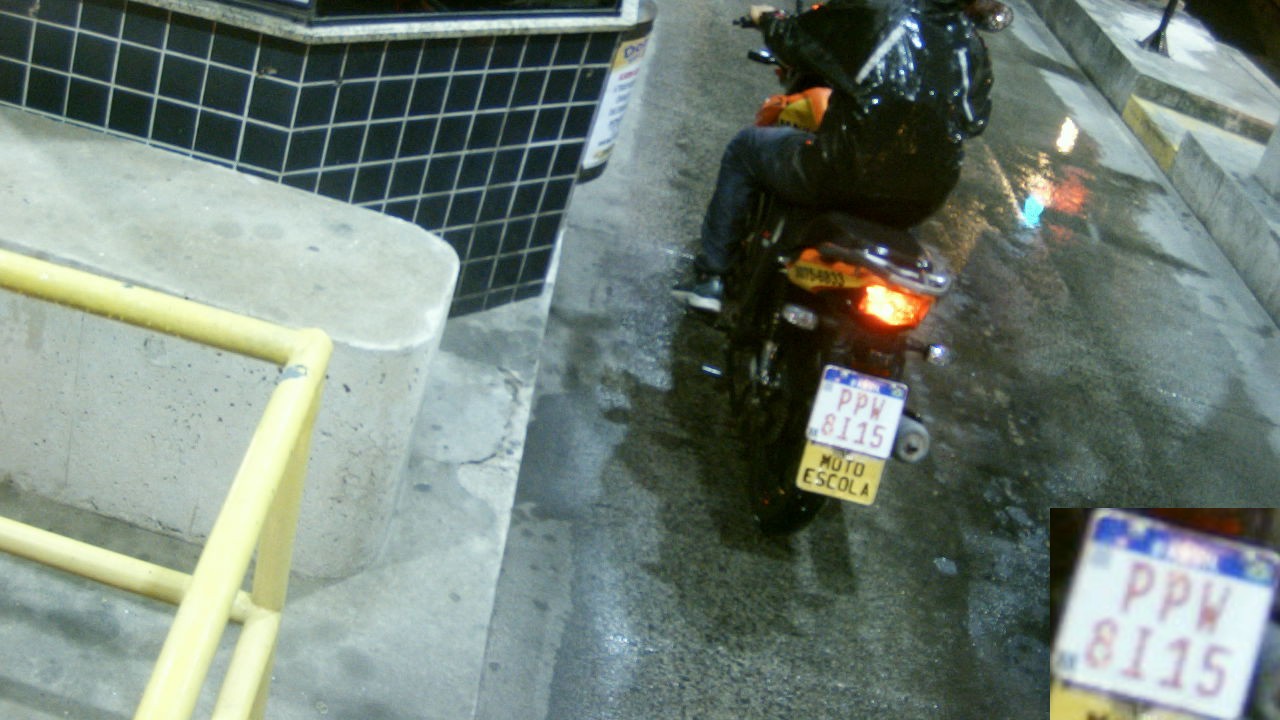}
    }

    \caption{Some images extracted from the \dataset dataset. 
    The first and second rows show images of cars and motorcycles, respectively, with Brazilian \glspl*{lp} (i.e., the standard used in Brazil before the adoption of the Mercosur standard).
    The third and fourth rows show images of cars and motorcycles, respectively, with Mercosur \glspl*{lp}. We show a zoomed-in version of the vehicle's \gls*{lp} in the lower right region of the images in the last column for better viewing of the \gls*{lp}~layouts.
    }
    \label{fig:samples-dataset}
\end{figure*}

\section{\uppercase{RodoSol-ALPR dataset}}
\label{sec:dataset}

The \dataset dataset contains $\numimages$ images captured by static cameras located at pay tolls owned by the \gls*{rodosol} concessionaire~\citep{rodosol} (hence the name of the dataset), which operates $\numkm$ kilometers of a highway (ES-060) in the Brazilian state of Esp\'{\i}rito Santo.

As can be seen in Figure~\ref{fig:samples-dataset}, there are images of different types of vehicles (e.g., cars, motorcycles, buses and trucks), captured during the day and night, from distinct lanes, on clear and rainy days, and the distance from the vehicle to the camera varies slightly.
All images have a resolution of $1{,}280 \times 720$~pixels. 

An important feature of the proposed dataset is that it has images of two different \gls*{lp} layouts: Brazilian and Mercosur.
To maintain consistency with previous works~\citep{izidio2020embedded,oliveira2021vehicle,silva2022flexible}, we refer to ``Brazilian'' as the standard used in Brazil before the adoption of the Mercosur standard.
All Brazilian \glspl*{lp} consist of three letters followed by four digits, while the initial pattern adopted in Brazil for Mercosur \glspl*{lp} consists of three letters, one digit, one letter and two digits, in that order.
In both layouts, car \glspl*{lp} have seven characters arranged in one row, whereas motorcycle \glspl*{lp} have three characters in one row and four characters in another.
Even though these \gls*{lp} layouts are very similar in shape and size, there are considerable differences in their colors and characters'~fonts.

The $\numimages$ images are divided as follows: $5{,}000$ images of cars with Brazilian \glspl*{lp}; $5{,}000$ images of motorcycles with Brazilian \glspl*{lp}; $5{,}000$ images of cars with Mercosur \glspl*{lp}; and $5{,}000$ images of motorcycles with Mercosur \glspl*{lp}.
For the sake of simplicity of definitions, here ``car'' refers to any vehicle with four wheels or more (e.g., passenger cars, vans, buses, trucks, among others), while ``motorcycle'' refers to both motorcycles and motorized tricycles.
As far as we know, \dataset is the public dataset for \gls*{alpr} with the highest number of motorcycle~images.

We randomly split the \dataset dataset as follows: $8{,}000$ images for training; $8{,}000$ images for testing; and $4{,}000$ images for validation, following the split protocol (i.e.,~$40$\%/$40$\%/$20$\%) adopted in the \ssigsegplate~\citep{goncalves2016benchmark} and \ufpralpr~\citep{laroca2018robust} datasets.
We preserved the percentage of samples for each vehicle type and \gls*{lp} layout; for example, there are $2{,}000$ images of cars with Brazilian \glspl*{lp} in each of the training and test sets, and $1{,}000$ images in the validation one.
For reproducibility purposes, the subsets generated are explicitly available along with the proposed~dataset.

Every image has the following information available in a text file: the vehicle's type (car or motorcycle), the \gls*{lp}'s layout (Brazilian or Mercosul), its text (e.g., ABC-1234), and the position~($x$,~$y$) of each of its four corners.
We labeled the corners instead of just the \gls*{lp} bounding box to enable the training of methods that explore \gls*{lp} rectification, as well as the application of a wider range of data augmentation techniques.

\begin{figure}[!htb]
    \centering
    \includegraphics[width=0.9\linewidth]{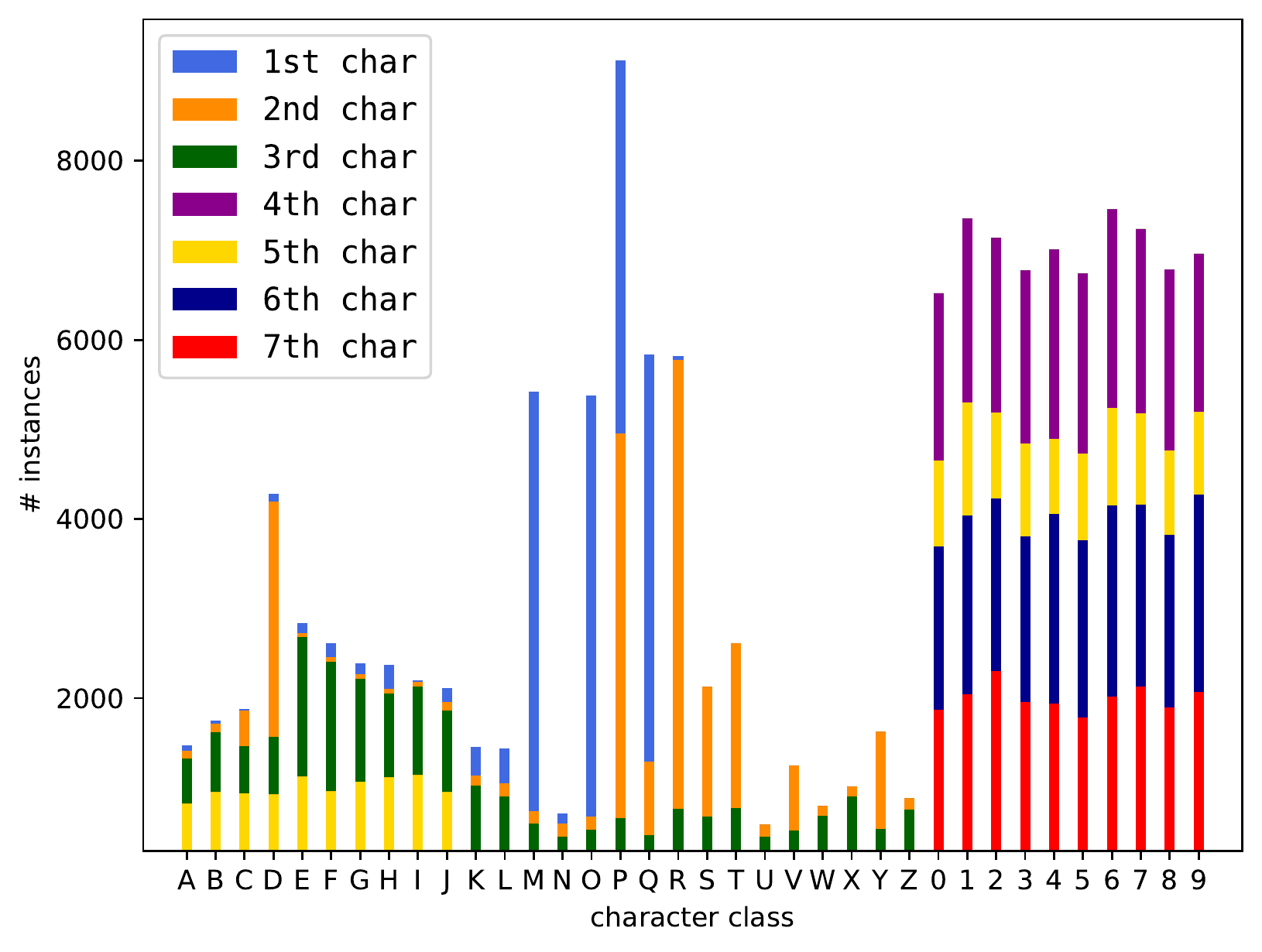} \,
    
    \vspace{-1.25mm}
    
    \caption{The distribution of character classes in the \dataset dataset. Observe that there is a significant imbalance in the distribution of the letters (due to \gls*{lp} allocation policies), whereas the digits are well balanced.}
    \label{fig:frequency}
\end{figure}

The datasets for \gls*{alpr} are generally very unbalanced in terms of character classes due to \gls*{lp} allocation policies~\citep{zhang2021robust_attentional}.
In Brazil, for example, one letter can appear much more often than others according to the state in which the \gls*{lp} was issued~\citep{goncalves2018realtime,laroca2018robust}.
This information must be taken into account when training recognition models in order to avoid undesirable biases --~this is usually done through data augmentation techniques~\citep{zhang2021robust_attentional,hasnat2021robust}; for example, a network trained exclusively in our dataset may learn to always classify the first character as `P' in cases where it should be `B' or~`R' since it appears much more often in this position than these two characters (see Figure~\ref{fig:frequency}).

Regarding privacy concerns related to our dataset, we remark that in Brazil the \glspl*{lp} are related to the respective vehicles, i.e., no public information is available about the vehicle drivers/owners~\citep{placa_veiculo_planalto,oliveira2021vehicle}.
Moreover, all human faces (e.g., drivers or \gls*{rodosol}'s employees) were manually redacted (i.e., blurred) in each~image.
\section{\uppercase{Experiments}}
\label{sec:experiments}

In this section, we describe the setup adopted in our experiments. 
We first list the models we implemented for our assessments, explaining why they were chosen and not others.
Afterward, we provide the implementation details, for example, which framework was used to train/test each model and the respective hyperparameters.
We then present and briefly describe the datasets used in our experiments, as well as the data augmentation techniques explored to avoid overfitting.
Lastly, we detail the evaluation protocols adopted by us, that is, which images from each dataset were used for training or testing in each experiment, and how we evaluate the performance of each~method.

\subsection{Methods}
\label{sec:experiments-methods}

In this work, we evaluate $\numbaselines$ \gls*{ocr} models for \gls*{lp} recognition:
\rare~\citep{shi2016robust}, \rtwoam~\citep{lee2016recursive}, \starnet~\citep{liu2016starnet}, \crnn~\citep{shi2017endtoend}, \grcnn~\citep{wang2017deep}, \holistic~\citep{spanhel2017holistic}, \multitaskgabriel~\citep{goncalves2019multitask}, \rosetta~\citep{borisyuk2018rosetta}, \trba~\citep{baek2019what}, \crnet~\citep{silva2020realtime}, \fastocr~\citep{laroca2021towards}, and \vitstrbase~\citep{atienza2021vitstr}. 
Table~\ref{tab:models} presents an overview of these methods, listing the original \gls*{ocr} application for which they were designed as well as the framework we used to train and evaluate~them.

\begin{table}[!htb]
\centering
\caption{\gls*{ocr} models explored in our experiments.}
\label{tab:models}

\vspace{-0.75mm}

\resizebox{0.99\linewidth}{!}{%
\begin{tabular}{@{}lllc@{}}
\toprule
\multicolumn{3}{c}{Model}                            & Original Application       \\ \midrule
\multicolumn{4}{l}{Framework: PyTorch\footnotemark[3]} 
\\
& & \rtwoam~\citep{lee2016recursive}                             & Scene Text Recognition       \\
& &  \rare~\citep{shi2016robust}                            & Scene Text Recognition       \\
& & \starnet~\citep{liu2016starnet}                         & Scene Text Recognition       \\
& & \crnn~\citep{shi2017endtoend}                             & Scene Text Recognition       \\
& & \grcnn~\citep{wang2017deep}                            & Scene Text Recognition       \\
& & \rosetta~\citep{borisyuk2018rosetta}                          & Scene Text Recognition       \\
& & \trba~\citep{baek2019what}                            & Scene Text Recognition       \\
& & \vitstrbase~\citep{atienza2021vitstr}                      & Scene Text Recognition      \\ \midrule
\multicolumn{4}{l}{Framework: Keras\footnotemark[4]} 
\\
& & \holistic~\citep{spanhel2017holistic} & License Plate Recognition     \\ 
& & \multitaskgabriel~\citep{goncalves2019multitask}                        & License Plate Recognition      \\ \midrule
\multicolumn{4}{l}{Framework: Darknet\footnotemark[5]} 
\\
& & \crnet~\citep{silva2020realtime}                           & License Plate Recognition  \\   
& & \fastocr~\citep{laroca2021towards}                        & Image-based Meter Reading               \\ \bottomrule
\end{tabular}%
}
\end{table}

These models were chosen/implemented by us for two main reasons: 
(i)~they have been employed for \gls*{ocr} tasks with promising/impressive results~\citep{baek2019what,atienza2021vitstr,laroca2021towards}, and (ii)~we believe we have the necessary knowledge to train/adjust them in the best possible way in order to ensure fairness in our experiments, as the authors provided enough details about the  architectures used, and also because we designed/employed similar networks in previous works (even the same ones in some cases)~\citep{goncalves2018realtime,goncalves2019multitask,laroca2019convolutional,laroca2021towards}.
Note that we are not aware of any work in the \gls*{alpr} literature where so many recognition models were explored in the~experiments.
\footnotetext[3]{\url{https://github.com/roatienza/deep-text-recognition-benchmark/}}
\footnotetext[4]{\url{https://keras.io/}}
\footnotetext[5]{\url{https://github.com/AlexeyAB/darknet/}}
\setcounter{footnote}{5}

The CR-NET and Fast-OCR models are based on the YOLO object detector~\citep{redmon2016yolo}.
Thus, they are trained to predict $35$ classes (0-9, A-Z, where `O' and `0' are detected/recognized jointly) using the bounding box of each \gls*{lp} character as input.
Although these methods have been attaining impressive results, they require laborious data annotations, i.e., each character's bounding box needs to be labeled for training them~\citep{wang2022rethinking}.
All the other $10$ models, on the other hand, output the \gls*{lp} characters in a segmentation-free manner, i.e., they predict  the characters holistically from the \gls*{lp} region without the need to detect/segment~them.
According to previous works~\citep{goncalves2018realtime,atienza2021vitstr,hasnat2021robust}, the generalizability of such segmentation-free models tends to improve significantly through the use of data augmentation.

\subsection{Setup}
\label{sec:experiments-setup}

All experiments were carried out on a computer with an AMD Ryzen Threadripper $1920$X $3.5$GHz CPU, $96$~GB of RAM ($2133$ MHz), HDD $7200$ RPM, and an NVIDIA Quadro RTX~$8000$ GPU~($48$~GB). 

Although run-time analysis is considered a critical factor in the ALPR literature~\citep{lubna2021automatic}, we consider such analysis beyond the scope of this work since we used different frameworks to implement the recognition models and there are probably differences in implementation and optimization between them --~we implemented each method using either the framework where it was originally implemented or well-known public repositories.
For example, the YOLO-based models were implemented using Darknet\footnotemark[5] while the models originally proposed for scene text recognition were trained and evaluated using a fork\footnotemark[3] of the open source repository of Clova AI Research (PyTorch) used to record the $1$st place of ICDAR2013 focused scene text and ICDAR2019~ArT, and $3$rd place of ICDAR2017 COCO-Text and ICDAR2019 ReCTS~(task1)~\citep{baek2019what}.

For completeness, below we list the hyperparameters used in each framework for training the \gls*{ocr} models; we remark that these hyperparameters were defined based on previous works fas well as on experiments performed in the validation set.
In Darknet, we employed the following parameters:~\gls*{sgd} optimizer, $90$K iterations (max batches), batch size~=~$64$, and learning rate~=~[$10$\textsuperscript{-$3$},~$10$\textsuperscript{-$4$},~$10$\textsuperscript{-$5$}] with decay steps at $30$K and $60$K~iterations.
In Keras, we used the Adam optimizer, initial learning rate~=~$10$\textsuperscript{-$3$} (with \textit{ReduceLROnPlateau}'s patience = $5$ and factor~=~$10$\textsuperscript{-$1$}), batch size~=~$64$, max epochs~=~$100$, and patience~=~$11$ (patience refers to the number of epochs with no improvement after which training is stopped).
In PyTorch, we adopted the following parameters: Adadelta optimizer, whose decay rate is set to $\rho= 0.99$,  $300$K iterations, and batch size~=~$128$.

\subsection{Datasets}
\label{sec:experiments-datasets}

Our experiments were conducted on images from the \dataset dataset and eight publicly available datasets that are often employed to benchmark \gls*{alpr} algorithms:
\caltech~\citep{caltech}, \englishlp~\citep{englishlp}, \stills~\citep{ucsd}, \chineselp~\citep{zhou2012principal}, \aolp~\citep{hsu2013application}, \openalpreu~\citep{openalpreu}, \ssigsegplate~\citep{goncalves2016benchmark}, \ufpralpr~\citep{laroca2018robust}.
Table~\ref{tab:experiments:overview_datasets} shows an overview of these datasets.
They were introduced over the last 22 years and have considerable diversity in terms of the number of images, acquisition settings, image resolution, and \gls*{lp} layouts.
As far as we know, there is no other work in the \gls*{alpr} literature where experiments were carried out on images from so many public~datasets.

\begin{table}[!htb]
\centering
\caption{The datasets used in our experiments. In this work, the ``Chinese'' layout refers to \glspl*{lp} of vehicles registered in mainland China, while the ``Taiwanese'' layout refers to \glspl*{lp} of vehicles registered in the Taiwan region.%
}
\label{tab:experiments:overview_datasets}

\vspace{-1mm}

\resizebox{0.95\columnwidth}{!}{ 
\begin{tabular}{@{}ccccc@{}}
\toprule
\textbf{Dataset} & \textbf{Year} & \textbf{Images} & \textbf{Resolution} & \textbf{LP Layout} \\ \midrule
\caltech & $1999$ & $126$ & $896\times592$ & American \\
\englishlp & $2003$ & $509$ & $640\times480$ & European \\
\stills & $2005$ & $291$ & $640\times480$ & American \\
\chineselp & $2012$ & $411$ & Various & Chinese \\
\aolp & $2013$ & $2049$ & Various & Taiwanese \\
\openalpreu & $2016$ & $108$ & Various & European \\
\ssigsegplate & $2016$ & $2000$ & $1920\times1080$ & Brazilian \\
\ufpralpr & $2018$ & $4500$ & $1920\times1080$ & Brazilian \\ 
\dataset & $2022$ & $20000$ & $1280\times720$ & Brazilian/Mercosur \\ \bottomrule
\end{tabular}
}
\end{table}

\begin{figure}[!htb]
    \centering
    \captionsetup[subfigure]{captionskip=-0.25pt,font={scriptsize},justification=centering} 
    
    \resizebox{0.9\linewidth}{!}{
	\subfloat[][\caltech]{
    \includegraphics[height=4.5ex]{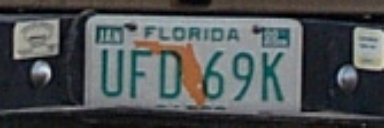}
    \includegraphics[height=4.5ex]{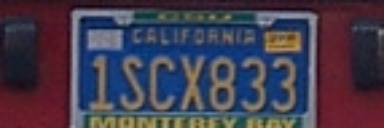}
    \includegraphics[height=4.5ex]{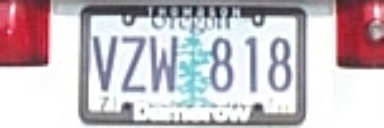}} \hspace{1mm} 
    }
    
    \vspace{0.75mm}
    
    \resizebox{0.9\linewidth}{!}{
	\subfloat[][\englishlp]{
    \includegraphics[height=4.5ex]{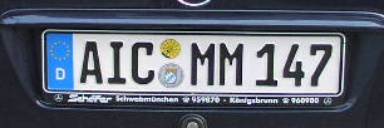}
    \includegraphics[height=4.5ex]{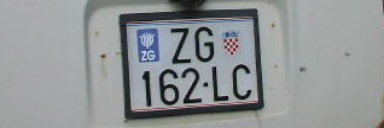}
    \includegraphics[height=4.5ex]{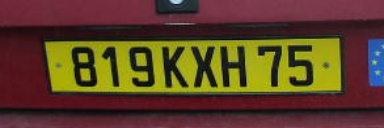}} \hspace{1mm} 
    }
    
    \vspace{0.75mm}
    
    \resizebox{0.9\linewidth}{!}{
	\subfloat[][\stills]{
    \includegraphics[height=4.5ex]{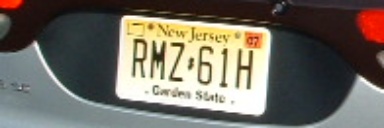}
    \includegraphics[height=4.5ex]{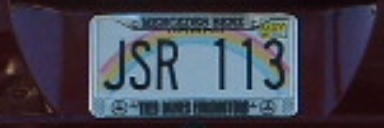}
    \includegraphics[height=4.5ex]{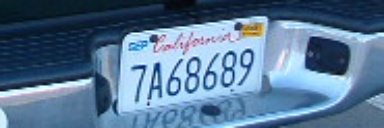}} \hspace{1mm} 
    }
    
    \vspace{0.75mm}
    
    \resizebox{0.9\linewidth}{!}{
	\subfloat[][\chineselp\label{fig:samples-public-datasets-chineselp}]{
    \includegraphics[height=4.5ex]{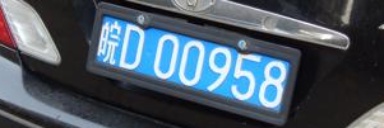}
    \includegraphics[height=4.5ex]{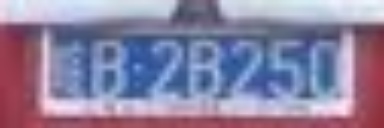}
    \includegraphics[height=4.5ex]{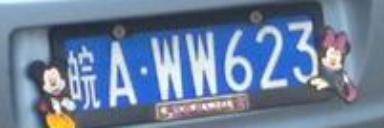}} \hspace{1mm} 
    }
    
    \vspace{0.75mm}
    
    \resizebox{0.9\linewidth}{!}{
	\subfloat[][\aolp]{
    \includegraphics[height=4.5ex]{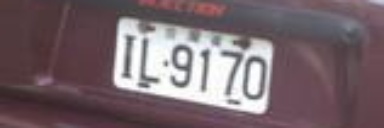}
    \includegraphics[height=4.5ex]{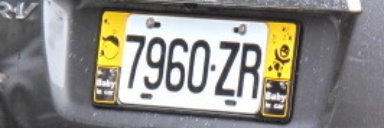}
    \includegraphics[height=4.5ex]{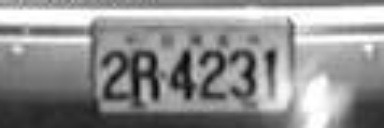}} \hspace{1mm} 
    }
    
    \vspace{0.75mm}
    
    \resizebox{0.9\linewidth}{!}{
	\subfloat[][\openalpreu]{
    \includegraphics[height=4.5ex]{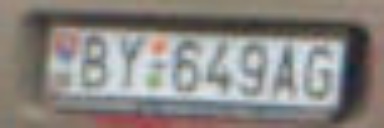}
    \includegraphics[height=4.5ex]{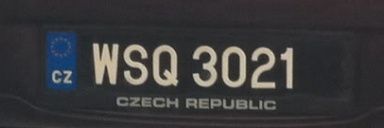}
    \includegraphics[height=4.5ex]{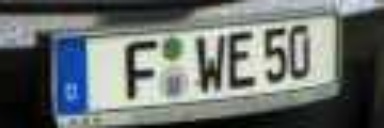}} \hspace{1mm} 
    }
    
    \vspace{0.75mm}
    
    \resizebox{0.9\linewidth}{!}{
	\subfloat[][\ssigsegplate]{
    \includegraphics[height=4.5ex]{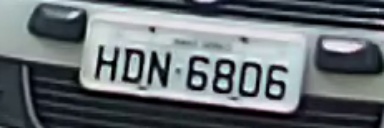}
    \includegraphics[height=4.5ex]{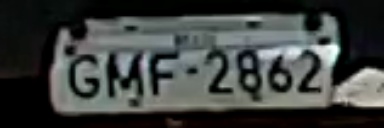}
    \includegraphics[height=4.5ex]{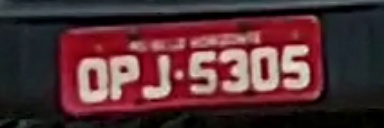}} \hspace{1mm} 
    }
    
    \vspace{0.75mm}
    
    \resizebox{0.9\linewidth}{!}{
	\subfloat[][\ufpralpr]{
    \includegraphics[height=4.5ex]{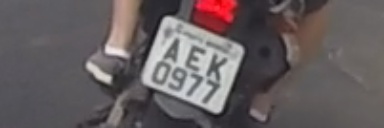}
    \includegraphics[height=4.5ex]{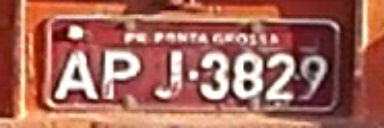}
    \includegraphics[height=4.5ex]{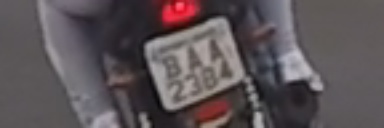}} \hspace{1mm} 
    }

    \vspace{-0.25mm}
    
    \caption{Some \gls*{lp} images from the public datasets used in our experimental evaluation.
    We show some \gls*{lp} images from the \dataset dataset in the last column of Fig~\ref{fig:samples-dataset}.
    }
    \label{fig:samples-public-datasets}
\end{figure}

Figure~\ref{fig:samples-public-datasets} shows the diversity of the chosen datasets in terms of \gls*{lp} layouts.
It is clear that even \glspl*{lp} from the same country can be quite different, e.g., the \caltech and \stills datasets were collected in the same region (California, United States), but they have images of \glspl*{lp} with significant differences in terms of colors, aspect ratios, backgrounds, and the number of characters.
It can also be observed that some datasets have \glspl*{lp} with two rows of characters and that the \glspl*{lp} may be tilted or have low resolution due to camera quality or vehicle-to-camera~distance.

In order to eliminate biases from the public datasets, we also used $772$ images from the internet --~those labeled and provided by \cite{laroca2021efficient}~-- to train all models.
These images include $257$ American \glspl*{lp}, $347$ Chinese \glspl*{lp}, and $178$ European \glspl*{lp}.
We chose not to use two datasets introduced recently: \karplate~\citep{henry2020multinational} and \ccpd~\citep{xu2018towards}.
The former cannot currently be downloaded due to legal problems.
The latter, although already available, was not employed for two main reasons: (i)~it contains highly compressed images, which significantly compromises the readability of the \glspl*{lp}~\citep{silva2022flexible};
and (ii)~it has some large errors in the corners' annotations~\citep{meng2020accelerating} --~this was somewhat expected since the corners were labeled automatically using RPnet~\citep{xu2018towards}. 
Additionally, we could not download the \clpd dataset~\citep{zhang2021robust_attentional}, as the authors made it available exclusively through a Chinese website where registration --~using a Chinese phone number or identity document~-- is required (we contacted the authors requesting an alternative link to download the dataset, but have not received a response so~far).

\subsubsection{Data Augmentation}

\begin{figure}[!htb]
    \centering
    
    \resizebox{0.95\linewidth}{!}{ %
    \includegraphics[height=4.5ex]{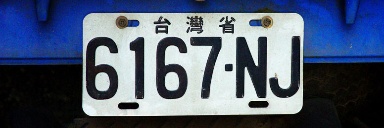} \hspace{-1.25mm}
    \includegraphics[height=4.5ex]{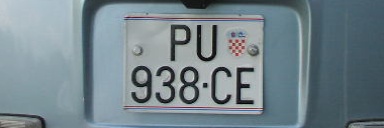} \hspace{-1.25mm}
    \includegraphics[height=4.5ex]{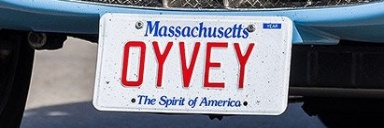}
    } %
    
    \vspace{0.1mm}
    
    \resizebox{0.95\linewidth}{!}{ %
    \includegraphics[height=4.5ex]{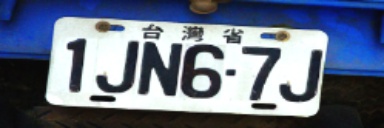} \hspace{-1.25mm} %
    \includegraphics[height=4.5ex]{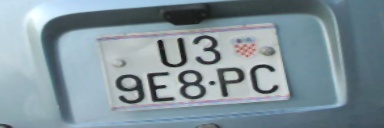} \hspace{-1.25mm}
    \includegraphics[height=4.5ex]{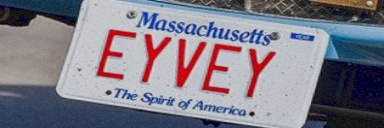} %
    } %
    
    \vspace{0.1mm}
    
    \resizebox{0.95\linewidth}{!}{ %
    \includegraphics[height=4.5ex]{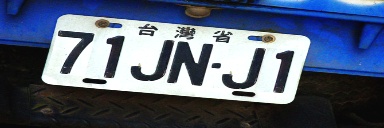}  \hspace{-1.25mm} %
    \includegraphics[height=4.5ex]{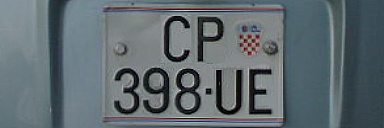} \hspace{-1.25mm}
    \includegraphics[height=4.5ex]{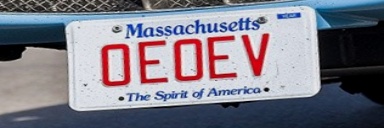} %
    } %
    
    \vspace{0.1mm}
    
    \resizebox{0.95\linewidth}{!}{ %
    \includegraphics[height=4.5ex]{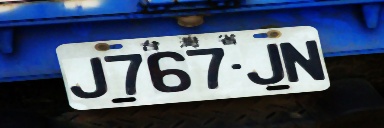} \hspace{-1.25mm} %
    \includegraphics[height=4.5ex]{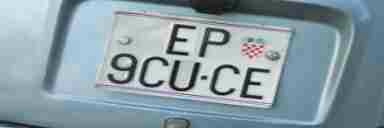} \hspace{-1.25mm}
    \includegraphics[height=4.5ex]{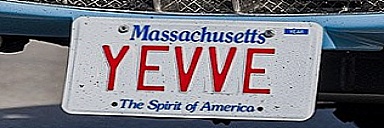} %
    } %
    
    \vspace{0.25mm}
    
    \caption{
    Illustration of the character permutation-based data augmentation technique~\citep{goncalves2018realtime} we adopted to avoid overfitting.
    The images in the first row are the originals, while the others were generated~automatically.}
    \label{fig:data-aug-preprocessing-ocr}
\end{figure}

As shown in Table~\ref{tab:experiments:overview_datasets}, two-thirds of the images used in our experiments are from the \dataset dataset.
In order to prevent overfitting, we initially balanced the number of images from different datasets through data augmentation techniques such as random cropping, random shadows, conversion to grayscale, and random perturbations of hue, saturation and brightness.
We used Albumentations~\citep{albumentations}, which is a well-known Python library for image augmentation, to apply these transformations.
Nevertheless, preliminary experiments showed that some of the recognition models were prone to predict only \gls*{lp} patterns that existed in the training set, as some patterns were being fed numerous times per epoch to the networks -- especially from small-scale datasets, where many images were created from a single original one.
Therefore, inspired by~\cite{goncalves2018realtime}, we also randomly permuted the position of the characters on each \gls*{lp} to eliminate such biases in the learning process (as illustrated in Figure~\ref{fig:data-aug-preprocessing-ocr}).
As the bounding box of each \gls*{lp} character is required to apply this data augmentation~technique --~these annotations are very time-consuming and laborious~-- we do not augment the training images from the RodoSol-ALPR dataset.
We believe this is not a significant problem as the proposed dataset is much larger than the~others.
The images from the other public datasets were augmented using the labels provided by~\cite{laroca2021efficient}.

In this process, we do not enforce the generated \glspl*{lp} to have the same arrangement of letters and digits of the original \glspl*{lp} so that the recognition models do not memorize specific patterns from different \gls*{lp} layouts.
For example, as described in Section~\ref{sec:dataset}, all Brazilian \glspl*{lp} consist of $3$ letters followed by $4$ digits, while Mercosur \glspl*{lp} have $3$ letters, $1$ digit, $1$ letter and $2$ digits, in that order.
Considering that \glspl*{lp} of these layouts are relatively similar (in size, shape, etc.), the segmentation-free networks would probably predict $3$ letters followed by $4$ digits for most Mercosur \gls*{lp} when holding the \dataset dataset out in a leave-one-dataset-out evaluation, as none of the other datasets have vehicles with Mercosur~\glspl*{lp}.

\subsection{Evaluation Protocols}
\label{sec:experiments-protocol}

In our experiments, we consider both traditional-split and leave-one-dataset-out protocols.
In the following subsections, we first describe them in~detail.
Then, we discuss how the performance evaluation is carried~out.

\subsubsection{Traditional Split}
\label{sec:experiments-protocol:traditional-split}

The traditional-split protocol assesses the ability of the models to perform well in seen scenarios, as each model is trained on the union of the training set images from all datasets and evaluated on the test set images from the respective datasets.
In recent works, the authors have chosen to train a single model on images from multiple datasets (instead of training a specific network for each dataset or \gls*{lp} layout as was commonly done in the past) so that the proposed models are robust for different scenarios with considerably less manual effort since their parameters are adjusted only once for all datasets~\citep{selmi2020delpdar,laroca2021efficient,silva2022flexible}.

For reproducibility, it is important to make clear how we divided the images from each of the datasets to train, validate and test the chosen models.
The \stills, \ssigsegplate, \ufpralpr and \dataset datasets were split according to the protocols defined by the respective authors, while the other datasets, which do not have well-defined evaluation protocols, were divided following previous works.
In summary, as in~\citep{xiang2019lightweight,henry2020multinational}, the \caltech dataset was randomly split into $80$ images for training/validation and $46$ images for testing.
Following~\citep{panahi2017accurate,beratoglu2021vehicle}, the \englishlp dataset was randomly divided as follows: $80$\% of the images for training/validation and $20$\% for testing.
For the \chineselp dataset, we employed the same protocol as \cite{laroca2021efficient}: $40$\% of the images for training, $20$\% for validation and $40$\% for testing.
We split each of the three subsets of the \aolp dataset (i.e., AC, LE, and RP) into training and test sets with a $2$:$1$ ratio, following~\citep{xie2018new,liang2021egsanet}, with $20$\%
of the training images being used for validation.
Finally, as most works in the literature~\citep{masood2017sighthound,laroca2021efficient,silva2022flexible}, we used all the $108$ images from the \openalpreu dataset for~testing (this division has been considered as a mini leave-one-dataset-out evaluation in recent works).
Table~\ref{tab:results:overview_datasets_protocols} lists the exact number of images used for training, validating and testing the chosen~models.

\begin{table}[!htb]
\centering
\caption{An overview of the number of images from each dataset used for training, validation, and testing.}
\label{tab:results:overview_datasets_protocols}
\vspace{-0.75mm}
\resizebox{0.95\linewidth}{!}{
\begin{tabular}{@{}cccccc@{}}
\toprule
\textbf{Dataset} & \textbf{Training} & \textbf{Validation} & \textbf{Testing} & \textbf{Discarded} & \textbf{Total} \\ \midrule
\caltech & $61$ & $16$ & $46$ & $3$ & $126$\\
\englishlp & $326$ & $81$ & $102$ & $0$ & $509$ \\
\stills & $181$ & $39$ & $60$ & $11$ & $291$\\
\chineselp & $159$ & $79$ & $159$ & $14$ & $411$\\
\aolp & $1{,}093$ & $273$ & $683$ & $0$ & $2{,}049$\\
\openalpreu & $0$ & $0$ & $108$ & $0$ & $108$ \\
\ssigsegplate & $789$ & $407$ & $804$ & $0$ & $2{,}000$\\
\ufpralpr & $1{,}800$ & $900$ & $1{,}800$ & $0$ & $4{,}500$ \\
\dataset & $8{,}000$ & $4{,}000$ & $8{,}000$ & $0$ & $20{,}000$ \\\bottomrule
\end{tabular}
}
\end{table}

As also detailed in Table~\ref{tab:results:overview_datasets_protocols}, a few images ($0.01$\%) were discarded in our experiments because it is impossible to recognize the \gls*{lp}(s) on them due to occlusion, lighting or image acquisition problems\footnote{\scriptsize The list of discarded images can be found at \discarded}.
Such images were also discarded by~\cite{masood2017sighthound} and~\cite{laroca2021efficient}.

\subsubsection{Leave-one-dataset-out}
\label{sec:experiments-protocol:leave-one-dataset-out}

The leave-one-dataset-out protocol evaluates the generalization performance of the trained models by testing them on the test set of an unseen dataset; that is, no images from that dataset are available during training.
For each experiment, we hold out the test set of one dataset as the unseen data, and train every model on all images from the other datasets.
As an example, if \aolp's test set is the current unseen data, the models are trained on all images from \caltech, \englishlp, \stills, \chineselp, \openalpreu, \ssigsegplate, \ufpralpr and \dataset, in addition to the images taken from the internet and provided by~\cite{laroca2021efficient}.

We evaluate the models only on the test set images from each unseen dataset, rather than including the training and validation images in the evaluation, so that the results achieved by each model on a given dataset are fully comparable with those achieved by the same model under the traditional-split~protocol.

\subsubsection{Performance Evaluation}
\label{sec:experiments-protocol:performance-evaluation}

As mentioned in Section~\ref{sec:introduction}, in our experiments, the \glspl*{lp} fed to the recognition models were detected using YOLOv4~\citep{bochkovskiy2020yolov4} --~with an input size of $672\times416$ pixels~-- rather than cropped directly from the ground truth. 
This procedure was adopted to better simulate real-world scenarios, as the \glspl*{lp} will not always be detected perfectly, and certain \gls*{ocr} models are not as robust in cases where the region of interest has not been detected so precisely~\citep{goncalves2018realtime}.
We employed the YOLOv4 model for this task because impressive results are consistently being reported in the \gls*{alpr} context through YOLO-based models~\citep{weihong2020research}.
Indeed, as detailed in Section~\ref{sec:results}, YOLOv4 reached an average recall rate above $99.5$\% in our experiments (we considered as correct the detections with \gls*{iou} $\ge0.5$ with the ground~truth).

For each experiment, we report the number of correctly recognized \glspl*{lp} divided by the number of \glspl*{lp} in the test set.
A correctly recognized \gls*{lp} means that all characters on the \gls*{lp} were correctly recognized, as a single incorrectly recognized character can result in the vehicle being incorrectly~identified.

Note that the first character in Chinese \glspl*{lp} is a Chinese character that represents the province in which the vehicle is affiliated~\citep{xu2018towards,zhang2021robust_attentional}.
Even though Chinese \glspl*{lp} are used in our experiments (see Figure~\ref{fig:samples-public-datasets-chineselp}), the evaluated models were not trained/adjusted to recognize Chinese characters; that is, only digits and English letters are considered.
This same procedure was adopted in previous works~\citep{li2019toward,selmi2020delpdar,laroca2021efficient} for several reasons, including scope reduction and the fact that it is not trivial for non-Chinese speakers to analyze the different Chinese characters in order to make an accurate error analysis or to choose which data augmentation techniques to~explore.
Following~\cite{li2019toward}, we denoted all Chinese characters as a single class~`*' in our experiments.
According to our results, the recognition models learned well the difference between Chinese characters and others --~i.e., digits and English letters~-- and this procedure did not affect the recognition rates~obtained.
\begin{table*}[!htb]
\centering
\setlength{\tabcolsep}{7pt}
\caption{Recall rates obtained by YOLOv4 in the \gls*{lp} detection stage.}
\label{tab:results-lp-detection}

\vspace{-1mm}

\resizebox{0.99\textwidth}{!}{%
\begin{tabular}{@{}lcccccccccc@{}}
\toprule
\diagbox[trim=l,innerrightsep=28.5pt]{Approach}{Test set}    & \multicolumn{1}{c}{\begin{tabular}[c]{@{}c@{}}\caltech\\\# $46$\phantom{\#}\end{tabular}} & \multicolumn{1}{c}{\begin{tabular}[c]{@{}c@{}}\englishlp\\\# $102$\phantom{\#}\end{tabular}} & \multicolumn{1}{c}{\begin{tabular}[c]{@{}c@{}}\stills\\\# $60$\phantom{\#}\end{tabular}} & \multicolumn{1}{c}{\begin{tabular}[c]{@{}c@{}}\chineselp\\\# $159$\phantom{\#}\end{tabular}} & \multicolumn{1}{c}{\begin{tabular}[c]{@{}c@{}}\aolp\\\# $683$\phantom{\#}\end{tabular}}    & \multicolumn{1}{c}{\begin{tabular}[c]{@{}c@{}}\openalpreu\\\# $108$\phantom{\#}\end{tabular}} & \multicolumn{1}{c}{\begin{tabular}[c]{@{}c@{}}\ssigsegplate\\\# $804$\phantom{\#}\end{tabular}} & \multicolumn{1}{c}{\begin{tabular}[c]{@{}c@{}}\ufpralpr\\\# $1{,}800$\phantom{\#}\end{tabular}} & \multicolumn{1}{c}{\begin{tabular}[c]{@{}c@{}}\dataset\\\# $8{,}000$\phantom{\#}\end{tabular}} & Average \\ \midrule
YOLOv4 (traditional-split)        & $100.0$\%      & $100.0$\%   & $100.0$\%     & $100.0$\%   & $100.0$\%  & $100.0$\%     & $\phantom{0}99.9$\%       & $99.1$\%       & $100.0$\%      & $99.9$\% \\
YOLOv4 (leave-one-dataset-out)        & $100.0$\%      & $100.0$\%   & $100.0$\%     & $100.0$\%   & $\phantom{0}99.9$\%  & $\phantom{0}99.1$\%     & $100.0$\%       & $96.8$\%       & $\phantom{0}99.6$\%      & $99.5$\% \\ \bottomrule
\end{tabular}%
}
\end{table*}
\begin{table*}[!htb]
\centering
\setlength{\tabcolsep}{7pt}
\caption{Recognition rates obtained by all models under the \ul{traditional-split} protocol, which assesses the ability of the models to perform well in seen scenarios.
Each model (rows) was trained once on the union of the training set images from all datasets and evaluated on the respective test sets (columns).
The best recognition rate achieved in each dataset is shown in bold.}
\label{tab:results-traditional-split}

\vspace{-1mm}

\resizebox{0.99\textwidth}{!}{%
\begin{tabular}{@{}lcccccccccc@{}}
\toprule
\diagbox[trim=l,innerrightsep=28.5pt]{Approach}{Test set}    & \multicolumn{1}{c}{\begin{tabular}[c]{@{}c@{}}\caltech\\\# $46$\phantom{\#}\end{tabular}} & \multicolumn{1}{c}{\begin{tabular}[c]{@{}c@{}}\englishlp\\\# $102$\phantom{\#}\end{tabular}} & \multicolumn{1}{c}{\begin{tabular}[c]{@{}c@{}}\stills\\\# $60$\phantom{\#}\end{tabular}} & \multicolumn{1}{c}{\begin{tabular}[c]{@{}c@{}}\chineselp\\\# $159$\phantom{\#}\end{tabular}} & \multicolumn{1}{c}{\begin{tabular}[c]{@{}c@{}}\aolp\\\# $683$\phantom{\#}\end{tabular}}    & \multicolumn{1}{c}{\begin{tabular}[c]{@{}c@{}}\openalpreu\\\# $108$\phantom{\#}\end{tabular}} & \multicolumn{1}{c}{\begin{tabular}[c]{@{}c@{}}\ssigsegplate\\\# $804$\phantom{\#}\end{tabular}} & \multicolumn{1}{c}{\begin{tabular}[c]{@{}c@{}}\ufpralpr\\\# $1{,}800$\phantom{\#}\end{tabular}} & \multicolumn{1}{c}{\begin{tabular}[c]{@{}c@{}}\dataset\\\# $8{,}000$\phantom{\#}\end{tabular}} & Average \\ \midrule
\crnet~\citep{silva2020realtime}    & $\textbf{95.7}$\textbf{\%}      & $92.2$\%   & $\textbf{100.0}$\textbf{\%}     & $96.9$\%   & $97.7$\% & $\textbf{97.2}$\textbf{\%}     & $97.1$\%       & $\textbf{78.3}$\textbf{\%}       & \phantom{$^\ddagger$}$55.8$\%$^\ddagger$      & $\textbf{90.1}$\textbf{\%} \\
\crnn~\citep{shi2017endtoend}        & $87.0$\%      & $81.4$\%   & $88.3$\%     & $88.2$\%   & $87.6$\%  & $89.8$\%     & $93.4$\%       & $64.9$\%       & $48.2$\%      & $81.0$\% \\
\fastocr~\citep{laroca2021towards}    & $93.5$\%      & $81.4$\%   & $95.0$\%     & $85.1$\%   & $95.8$\% & $91.7$\%     & $87.1$\%       & $65.9$\%       & \phantom{$^\ddagger$}$49.7$\%$^\ddagger$      & $82.8$\% \\
\grcnn~\citep{wang2017deep}       & $93.5$\%      & $87.3$\%   & $91.7$\%     & $84.5$\%   & $85.9$\% & $87.0$\%     & $94.3$\%       & $63.3$\%      & $48.4$\%      & $81.7$\% \\
\holistic~\citep{spanhel2017holistic}     & $89.1$\%      & $68.6$\%   & $88.3$\%     & $90.7$\%   & $86.3$\% & $78.7$\%     & $94.8$\%       & $70.3$\%       & $49.0$\%      & $79.5$\% \\
\multitaskgabriel~\citep{goncalves2019multitask}        & $87.0$\%      & $62.7$\%   & $85.0$\%     & $86.3$\%   & $84.7$\%  & $66.7$\%     & $93.0$\%       & $65.3$\%       & $49.1$\%      & $75.5$\% \\
\rtwoam~\citep{lee2016recursive}        & $84.8$\%      & $70.6$\%   & $81.7$\%     & $87.0$\%   & $83.1$\%  & $63.9$\%     & $92.0$\%       & $66.9$\%       & $48.6$\%      & $75.4$\% \\
\rare~\citep{shi2016robust}        & $91.3$\%      & $84.3$\%   & $90.0$\%     & $95.7$\%   & $93.4$\% & $91.7$\%     & $93.7$\%       & $69.0$\%       & $51.6$\%      & $84.5$\% \\
\rosetta~\citep{borisyuk2018rosetta}     & $87.0$\%      & $75.5$\%   & $81.7$\%     & $90.1$\%   & $83.7$\% & $81.5$\%     & $94.3$\%       & $63.9$\%       & $48.7$\%      & $78.5$\% \\
\starnet~\citep{liu2016starnet}    & $\textbf{95.7}$\textbf{\%}      & $\textbf{93.1}$\textbf{\%}   & $96.7$\%     & $96.9$\%   & $96.8$\% & $95.4$\%     & $96.1$\%       & $70.9$\%       & $51.8$\%      & $88.2$\% \\
\trba~\citep{baek2019what}        & $91.3$\%      & $87.3$\%   & $96.7$\%     & $96.9$\%   & $\textbf{99.0}$\textbf{\%} & $93.5$\%     & $\textbf{97.3}$\textbf{\%}       & $72.9$\%       & $\textbf{59.6}$\textbf{\%}      & $88.3$\% \\
\vitstrbase~\citep{atienza2021vitstr} & $84.8$\%      & $80.4$\%   & $90.0$\%     & $\textbf{99.4}$\textbf{\%}   & $95.6$\%  & $84.3$\%     & $96.1$\%       & $73.3$\%       & $49.3$\%      & $83.7$\% \\
\midrule
Average    & $90.0$\%      & $80.4$\%   & $90.4$\%     & $91.5$\%   & $90.8$\% & $85.1$\%     & $94.1$\%       & $68.7$\%       & $50.8$\%      & $\acctraditional$\% \\ \bottomrule \\[-2.2ex]
\multicolumn{11}{l}{\small $^{\ddagger}$Images from the \dataset dataset were not used for training the \crnet and \fastocr models, as each character’s bounding box needs to be labeled for training them (as detailed in Section~\ref{sec:experiments-methods}).}
\end{tabular}%
}
\end{table*}
\begin{table*}[!htb]
\centering
\setlength{\tabcolsep}{7pt}
\caption{Recognition rates obtained by all models under the \ul{leave-one-dataset-out} protocol, which assesses the generalization performance of the models by testing them on the test set of an unseen dataset.
For each dataset (columns), we trained the recognition models (rows) on all images from the other datasets.
The best recognition rates achieved are shown in bold.
}
\label{tab:results-leave-one-dataset-out}

\vspace{-1mm}

\resizebox{0.99\textwidth}{!}{%
\begin{tabular}{@{}lcccccccccc@{}}
\toprule
\diagbox[trim=l,innerrightsep=28.5pt]{Approach}{Test set}    & \multicolumn{1}{c}{\begin{tabular}[c]{@{}c@{}}\caltech\\\# $46$\phantom{\#}\end{tabular}} & \multicolumn{1}{c}{\begin{tabular}[c]{@{}c@{}}\englishlp\\\# $102$\phantom{\#}\end{tabular}} & \multicolumn{1}{c}{\begin{tabular}[c]{@{}c@{}}\stills\\\# $60$\phantom{\#}\end{tabular}} & \multicolumn{1}{c}{\begin{tabular}[c]{@{}c@{}}\chineselp\\\# $159$\phantom{\#}\end{tabular}} & \multicolumn{1}{c}{\begin{tabular}[c]{@{}c@{}}\aolp\\\# $683$\phantom{\#}\end{tabular}}    & \multicolumn{1}{c}{\begin{tabular}[c]{@{}c@{}}\openalpreu\\\# $108$\phantom{\#}\end{tabular}} & \multicolumn{1}{c}{\begin{tabular}[c]{@{}c@{}}\ssigsegplate\\\# $804$\phantom{\#}\end{tabular}} & \multicolumn{1}{c}{\begin{tabular}[c]{@{}c@{}}\ufpralpr\\\# $1{,}800$\phantom{\#}\end{tabular}} & \multicolumn{1}{c}{\begin{tabular}[c]{@{}c@{}}\dataset\\\# $8{,}000$\phantom{\#}\end{tabular}} & Average \\ \midrule
\crnet~\citep{silva2020realtime}   & $93.5$\%      & $\textbf{96.1}$\textbf{\%}   & $\textbf{96.7}$\textbf{\%}     & $88.2$\%   & $76.9$\% & $\textbf{96.3}$\textbf{\%}     & $94.7$\%       & $61.8$\%      & $45.4$\%      & $\textbf{83.3}$\textbf{\%} \\ 
\crnn~\citep{shi2017endtoend}        & $91.3$\%      & $62.7$\%   & $75.0$\%     & $76.4$\%   & $59.4$\% & $88.0$\%     & $91.3$\%       & $61.7$\%       & $38.8$\%      & $71.6$\% \\
\fastocr~\citep{laroca2021towards}     & $93.5$\%      & $91.2$\%   & $95.0$\%    & $90.1$\%   & $77.0$\% & $94.4$\%     & $91.2$\%       & $53.2$\%      & $\textbf{47.8}$\textbf{\%}      & $81.5$\% \\
\grcnn~\citep{wang2017deep}       & $\textbf{95.7}$\textbf{\%}      & $65.7$\%   & $90.0$\%     & $80.7$\%   & $53.9$\% & $88.9$\%     & $90.3$\%       & $60.8$\%       & $39.8$\%      & $74.0$\% \\
\holistic~\citep{spanhel2017holistic}        & $80.4$\%      & $40.2$\%   & $73.3$\%     & $81.4$\%   & $59.7$\%  & $83.3$\%     & $93.4$\%       & $61.8$\%      & $33.4$\%      & $67.4$\% \\
\multitaskgabriel~\citep{goncalves2019multitask}        & $82.6$\%     & $34.3$\%    & $66.7$\%      & $77.6\%$    & $50.8$\%   & $79.6$\%     & $89.9$\%       & $57.9$\%      & $44.8$\%      & $64.9$\% \\
\rtwoam~\citep{lee2016recursive}        & $89.1$\%      & $52.9$\%   & $66.7$\%     & $74.5$\%   & $52.5$\%  & $80.6$\%     & $93.5$\%       & $57.9$\%       & $40.7$\%      & $67.6$\% \\
\rare~\citep{shi2016robust}        & $84.8$\%      & $50.0$\%   & $85.0$\%     & $88.8$\%   & $62.9$\% & $91.7$\%     & $93.5$\%       & $71.3$\%       & $40.1$\%      & $74.2$\% \\
\rosetta~\citep{borisyuk2018rosetta}     & $89.1$\%      & $63.7$\%   & $68.3$\%     & $83.2$\%   & $51.1$\% & $81.5$\%     & $94.4$\%       & $61.8$\%        & $42.5$\%      & $70.6$\% \\
\starnet~\citep{liu2016starnet}    & $89.1$\%      & $80.4$\%   & $91.7$\%     & $\textbf{95.0}$\textbf{\%}   & $\textbf{79.3}$\textbf{\%} & $93.5$\%     & $94.0$\%       & $69.1$\%      & $43.6$\%      & $81.8$\% \\
\trba~\citep{baek2019what}        & $\textbf{95.7}$\textbf{\%}      & $66.7$\%   & $93.3$\%     & $\textbf{95.0}$\textbf{\%}   & $70.0$\% & $92.6$\%     & $96.9$\%       & $\textbf{73.2}$\textbf{\%}      & $42.6$\%      & $80.7$\% \\
\vitstrbase~\citep{atienza2021vitstr} & $89.1$\%      & $58.8$\%   & $90.0$\%     & $\textbf{95.0}$\textbf{\%}   & $59.2$\%  & $89.8$\%     & $\textbf{97.9}$\textbf{\%}       & $69.6$\%      & $41.7$\%      & $76.8$\% \\[2pt] \cdashline{1-11} \\[-7.5pt]
Average   & $89.5$\%      & $63.6$\%   & $82.6$\%     & $85.5$\%   & $62.7$\% & $88.3$\%     & $93.4$\%       & $63.3$\%      & $41.8$\%      & $\acclodo$\% \\ %
Average (traditional-split protocol)    & $90.0$\%      & $80.4$\%   & $90.4$\%     & $91.5$\%   & $90.8$\% & \phantom{$^\dagger$}$85.1$\%$^\dagger$     & $94.1$\%       & $68.7$\%       & $50.8$\%      & $\acctraditional$\% 
\\ \midrule
Sighthound~\citep{masood2017sighthound}  & $87.0$\%      & $94.1$\%   & $90.0$\%     & $84.5$\%   & $79.6$\% & $94.4$\%     & $79.2$\%       & $52.6$\%       & $51.0$\%      & $79.2$\% \\
OpenALPR~\citep{openalprapi}$^\ast$           & $95.7$\%      & $99.0$\%   & $96.7$\%     & $93.8$\%   & $81.1$\% & $99.1$\%     & $91.4$\%       & $87.8$\%       & $70.0$\%      & $90.5$\% \\ \bottomrule \\[-2.2ex]
\multicolumn{11}{l}{\small $^{\dagger}$Under the traditional-split protocol, no images from the \openalpreu dataset were used for training. This is the protocol commonly adopted in the literature~\citep{laroca2021efficient,silva2022flexible}.} \\
\multicolumn{11}{l}{\small $^{\ast}$OpenALPR contains specialized solutions for \glspl*{lp} from different regions and the user must enter the correct region before using its API. Hence, it was expected to achieve better results than the other methods.}
\end{tabular}%
}
\end{table*}

\section{\uppercase{Results and Discussion}}
\label{sec:results}

First, we report in Table~\ref{tab:results-lp-detection} the recall rates obtained by the YOLOv4 model in the \gls*{lp} detection stage.
As can be seen, it reached surprisingly good results in both protocols.
More specifically, recall rates above $99.9$\% were achieved in $14$ of the $18$ assessments. 
As in previous works~\citep{laroca2018robust,goncalves2018realtime,silva2020realtime}, the detection results are slightly worse for the \ufpralpr dataset due to its challenging nature, as (i)~it has images where the vehicles are considerably far from the camera;
(ii)~some of its frames have motion blur because the dataset was recorded in real-world scenarios where both the vehicle and the camera --~inside another vehicle~-- are moving;
and (iii)~it also contains images of motorcycles, where the backgrounds can be much more complicated due to different body configurations and mixtures with other background~scenes~\citep{hsu2015comparison}.

Considering the discussion above, we assert that deep models trained for \gls*{lp} detection on images from multiple datasets can be employed quite reliably on images from unseen datasets (i.e., leave-one-dataset-out protocol).
Of course, this may not hold true in extraordinary cases where the test set domain is very different from training ones, but this was not the case in our experimental evaluation carried out on images from nine datasets with different~characteristics.

Regarding the recognition stage, the results achieved by all models across all datasets on the traditional-split and leave-one-dataset-out protocols are shown in Table~\ref{tab:results-traditional-split} and Table~\ref{tab:results-leave-one-dataset-out}, respectively.
In Table~\ref{tab:results-leave-one-dataset-out}, we included the results obtained by Sighthound~\citep{masood2017sighthound} and OpenALPR~\citep{openalprapi}, which are two commercial systems frequently used as baselines in the \gls*{alpr} literature, since in principle they are trained on images from large-scale private datasets and not from the public datasets explored here (i.e., leave-one-dataset-out~protocol).

The first observation is that, as expected, the best results --~on average for all models~-- were attained when training and evaluating the models on different subsets from the same datasets (i.e., traditional-split protocol).
The only case where this did not occur was precisely in the \openalpreu dataset, where no images are used for training even under the traditional-split protocol (see Table~\ref{tab:results:overview_datasets_protocols}).
We kept this division for three main reasons: (i)~to better evaluate the recognition models on European \glspl*{lp}; (ii)~to maintain consistency with previous works~\citep{masood2017sighthound,laroca2021efficient,silva2022flexible}, which also used all images from that dataset for testing; and (iii)~to analyze how the models perform with more training data from other datasets, which in this case corresponds to the leave-one-dataset-out protocol since all images from the other datasets --~and not just the training set ones~-- are used for~training.
Although some studies have shown that the performance on the test set of a particular dataset often decreases when the training data is augmented with data from other datasets~\citep{torralba2011unbiased,khosla2012undoing}, the recognition rates reached in the \openalpreu dataset were higher with more training data from other datasets.
In the same way, \crnet performed better in the \englishlp dataset when using all images from the \openalpreu dataset for~training (both datasets contain images of European \glspl*{lp}). 

\begin{figure}[!htb]
    \centering
    \captionsetup[subfigure]{labelformat=empty,font={normalsize},captionskip=0.3pt}
    
    \resizebox{\linewidth}{!}{
    \subfloat[][\centering \phantom{\,} \resizebox{\adj}{!}{\textbf{LODO:}} \texttt{7615\wrong{B}G}\hspace{\textwidth} \phantom{\,}  \resizebox{\adj}{!}{\textbf{Trad.:}} \texttt{7615RG}]{
		\includegraphics[width=0.32\linewidth]{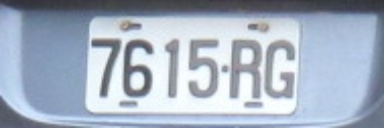}} \, %
	\subfloat[][\centering \phantom{\,} \resizebox{\adj}{!}{\textbf{LODO:}} \texttt{P\wrong{G}379\wrong{I}}\hspace{\textwidth} \phantom{\,} \resizebox{\adj}{!}{\textbf{Trad.:}} \texttt{P63791}]{
		\includegraphics[width=0.32\linewidth]{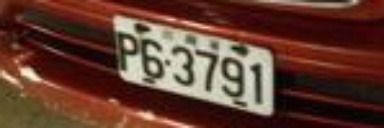}} \, %
    \subfloat[][\centering \phantom{\,} \resizebox{\adj}{!}{\textbf{LODO:}} \texttt{\wrong{0}X7655}\hspace{\textwidth} \phantom{\,}  \resizebox{\adj}{!}{\textbf{Trad.:}} \texttt{DX7655}]{
    		\includegraphics[width=0.32\linewidth]{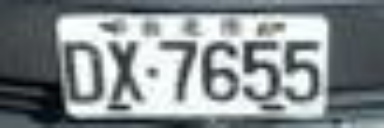}} \hspace{1.5mm}
	}
	
    \caption{The predictions obtained by \trba on three images of the \aolp dataset. In general, the errors (outlined in red) under the leave-one-dataset-out~(LODO) protocol did not occur in challenging cases (e.g., blurry or tilted images); therefore, they were probably caused by \emph{differences in the training and test images}. Trad.: traditional-split protocol.}
    \label{fig:samples-errors-aolp}
\end{figure}

The average recognition rate across all datasets decreased from $\acctraditional$\% under the traditional-split protocol to $\acclodo$\% under the leave-one-dataset-out protocol.
This drastic performance drop is accentuated by the poor results achieved on the \englishlp and \aolp datasets under the leave-one-dataset-out protocol.
For instance, the average recognition rate of $90.8$\% obtained in the \aolp dataset under the traditional-split protocol drops to $62.7$\% under the leave-one-dataset-out protocol.
These results caught us by surprise, as both datasets have been considered to be relatively simple due to the fact that recent works have reported recognition rates close to 97\% for the \englishlp and above 99\% for the \aolp dataset~\citep{henry2020multinational,laroca2021efficient,silva2022flexible,wang2022rethinking}.
According to our analysis, most of the recognition errors under the leave-one-dataset-out protocol occurred due to differences in the fonts of the \gls*{lp} characters in the training and test images, as well as because of specific patterns in the \gls*{lp} (e.g., a coat of arms between the \gls*{lp} characters or a straight line under them).
To better illustrate, Figure~\ref{fig:samples-errors-aolp} shows three \glspl*{lp} from the \aolp dataset where the \trba model, which performed best on that dataset, recognized at least one character incorrectly under the leave-one-dataset-out protocol but not under the traditional split.
Such analysis highlights the importance of performing cross-dataset experiments in the \gls*{alpr}~context.

The second observation is that, regardless of the evaluation protocol adopted, no recognition model achieved the best results in every single dataset we performed experiments on.
For instance, although the \crnet model obtained the best average recognition rates, corroborating the state-of-the-art results reported recently in \citep{laroca2021efficient,silva2022flexible}, it did not reach the best results in the \chineselp, \aolp, \ssigsegplate and \dataset datasets in either protocol.
These results emphasize the importance of carrying out experiments on multiple datasets, with \glspl*{lp} from different countries/regions, especially under the leave-one-dataset-out protocol because six different models obtained the best result in at least one~dataset.

The third observation is that the \dataset dataset proved very challenging since all the recognition models trained by us, as well as both commercial systems, failed to reach recognition rates above $70$\% on its test set images. 
The main reason for such disappointing results is the large number of motorcycle images, which are very challenging in nature (as discussed in Section~\ref{sec:related_work}).
For example, OpenALPR correctly recognized $3{,}772$ of the $4{,}000$ cars in the test set ($94.3$\%) and only $1{,}827$ of the $4{,}000$ motorcycles in the test set ($45.7$\%).
These results accentuate the importance of the proposed dataset for the accurate evaluation of \gls*{alpr} systems, as it avoids bias in the assessments by having the same number of ``easy'' (cars with single-row \glspl*{lp}) and ``difficult'' (motorcycles with two-row \glspl*{lp})~samples.

\begin{figure}[!htb]
    \centering
    \captionsetup[subfigure]{labelformat=empty,font={small},captionskip=0.25pt}
    
    \resizebox{\linewidth}{!}{
    \subfloat[][\centering \phantom{\,} \resizebox{\adjj}{!}{\phantom{AAAA}\textbf{\trba:}} \texttt{HLP\wrong{A}594}\hspace{\textwidth} \phantom{\,}  \resizebox{\adjj}{!}{\textbf{OpenALPR:}} \texttt{HLP4594}\hspace{\textwidth} \phantom{\,}  \resizebox{\adjj}{!}{\phantom{AAAAa}\textbf{GT:}} \texttt{HLP4594}]{
		\includegraphics[width=0.32\linewidth]{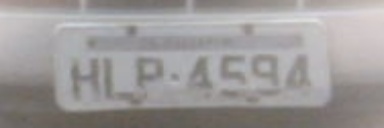}} \, %
	\subfloat[][\centering \phantom{\,} \resizebox{\adjj}{!}{\phantom{AAAA}\textbf{\trba:}} \texttt{PPY6\wrong{0}26}\hspace{\textwidth} \phantom{\,}  \resizebox{\adjj}{!}{\textbf{OpenALPR:}} \texttt{PPY6\wrong{0}26}\hspace{\textwidth} \phantom{\,}  \resizebox{\adjj}{!}{\phantom{AAAAa}\textbf{GT:}} \texttt{PPY6C26}]{
		\includegraphics[width=0.32\linewidth]{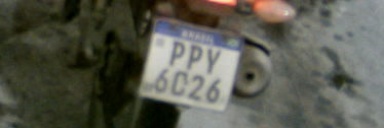}} \, %
    \subfloat[][\centering \phantom{\,} \resizebox{\adjj}{!}{\phantom{AAAA}\textbf{\trba:}} \texttt{QRE4E6\wrong{Z}}\hspace{\textwidth} \phantom{\,}  \resizebox{\adjj}{!}{\textbf{OpenALPR:}} \texttt{QRE4E62}\hspace{\textwidth} \phantom{\,}  \resizebox{\adjj}{!}{\phantom{AAAAa}\textbf{GT:}} \texttt{QRE4E62}]{
		\includegraphics[width=0.32\linewidth]{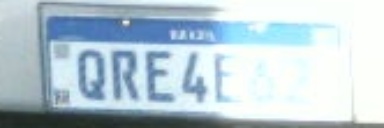}} \, %
	}
	
	\vspace{2mm}
	
	\resizebox{\linewidth}{!}{
    \subfloat[][\centering \phantom{\,} \resizebox{\adjj}{!}{\phantom{AAAA}\textbf{\trba:}} \texttt{QRG6D57}\hspace{\textwidth} \phantom{\,}  \resizebox{\adjj}{!}{\textbf{OpenALPR:}} \texttt{\wrong{-------}}\hspace{\textwidth} \phantom{\,}  \resizebox{\adjj}{!}{\phantom{AAAAa}\textbf{GT:}} \texttt{QRG6D57}]{
		\includegraphics[width=0.32\linewidth]{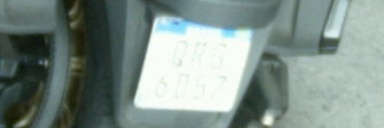}} \, %
	\subfloat[][\centering \phantom{\,} \resizebox{\adjj}{!}{\phantom{AAAA}\textbf{\trba:}} \texttt{O\wrong{O}M8060}\hspace{\textwidth} \phantom{\,}  \resizebox{\adjj}{!}{\textbf{OpenALPR:}} \texttt{O\wrong{O}M8060}\hspace{\textwidth} \phantom{\,}  \resizebox{\adjj}{!}{\phantom{AAAAa}\textbf{GT:}} \texttt{ODM8060}]{
		\includegraphics[width=0.32\linewidth]{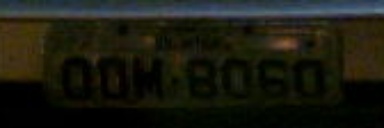}} \, %
    \subfloat[][\centering \phantom{\,} \resizebox{\adjj}{!}{\phantom{AAAA}\textbf{\trba:}} \texttt{MR\wrong{O}3095}\hspace{\textwidth} \phantom{\,}  \resizebox{\adjj}{!}{\textbf{OpenALPR:}} \texttt{MR\wrong{O}3095}\hspace{\textwidth} \phantom{\,}  \resizebox{\adjj}{!}{\phantom{AAAAa}\textbf{GT:}} \texttt{MRU3095}]{
		\includegraphics[width=0.32\linewidth]{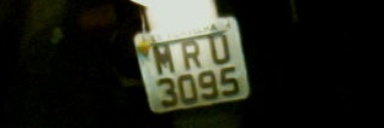}} \, %
	}
	
    \caption{Some \gls*{lp} images from \dataset along with the predictions returned by \trba and OpenALPR. Note that one character may become very similar to another due to factors such as blur, low/high exposure, rotations and occlusions.
    For correctness, we checked if the ground truth (GT) matched the vehicle make and model on the \gls*{denatran}~database.}
    \label{fig:samples-dataset-challenging}
\end{figure}

We also did not rule out challenging images when selecting the images for the creation of the dataset.
Figure~\ref{fig:samples-dataset-challenging} shows some of these images along with the predictions returned by \trba (traditional-split) and OpenALPR, which were the model trained by us and the commercial system that performed better on this dataset.
The results are in line with what was recently stated by~\cite{zhang2021robust_attentional}: that recognizing \glspl*{lp} in complex environments is still far from~satisfactory.

Lastly, it is important to highlight the number of experiments we carried out for this work.
We trained each of the $\numbaselines$ chosen \gls*{ocr} models $10$ times: once following the split protocols traditionally adopted in the literature (see Table~\ref{tab:results-traditional-split}) and nine for the leave-one-dataset-out evaluation (see Table~\ref{tab:results-leave-one-dataset-out}).
We remark that a single training process of some models (e.g., \trba and \vitstrbase) took several days to complete on an NVIDIA Quadro RTX $8000$ GPU.
In fact, we believe that this large number of necessary experiments is precisely what caused a leave-one-dataset-out evaluation to have not yet been performed in the~literature.

\section{\uppercase{Conclusions}}
\label{sec:conclusions}

As the performance of traditional-split \gls*{lp} recognition is rapidly improving, researchers should pay more attention to cross-dataset \gls*{lp} recognition since it better simulates real-world \gls*{alpr} applications, where new cameras are regularly being installed in new locations without existing systems being retrained every~time.

As a first step towards that direction, in this work we evaluated $\numbaselines$ \gls*{ocr} models for \gls*{lp} recognition on $9$ public datasets with a great variety in several aspects (e.g., acquisition settings, image resolution, and \gls*{lp} layouts).
We adopted a traditional-split \textit{versus} leave-one-dataset-out experimental setup to empirically assess the cross-dataset generalization of the chosen models.
It is noteworthy that we are not aware of any work in the \gls*{alpr} context where so many methods were implemented and compared or where so many datasets were explored in the~experiments.

As expected, the experimental results showed significant drops in performance for most datasets when training and testing the recognition models in a leave-one-dataset-out fashion.
The fact that very low recognition rates (around $63$\%) were reported in the \englishlp and \aolp datasets underscored the importance of carrying out cross-dataset experiments, as very high recognition rates (above $95$\% and $99$\%, respectively) are frequently achieved on these datasets under the traditional-split~protocol.

The importance of exploring various datasets in the evaluation was also demonstrated, as no model performed better than the others in all experiments.
It was quite unexpected for us that six different models reached the best result in at least one dataset under the leave-one-dataset-out protocol.
In this sense, we draw attention to the fact that most works in the literature used three or fewer datasets in the experiments, although this has been gradually changing in recent years~\citep{selmi2020delpdar,laroca2021efficient}.

We also introduced a publicly available dataset for \gls*{alpr} that, to the best of our knowledge, is the first to contain images of vehicles with Mercosur \glspl*{lp}.
We expect it will assist in developing new approaches for this \gls*{lp} layout and the fair comparison between methods proposed in different~works.
Additionally, the proposed dataset includes $10{,}000$ motorcycle images, being by far the largest in this regard.
\dataset has proved challenging in our experiments, as both the models trained by us and two commercial systems reached recognition rates below $70$\% on its test~set.

As future work, we plan to gather images from the internet to build a novel dataset for end-to-end \gls*{alpr} with images acquired in various countries/regions, by many different cameras, both static or mobile, with a well-defined evaluation protocol for both within- and cross-dataset \gls*{lp} detection and \gls*{lp} recognition.
In addition, we intend to leverage the potential of \glspl*{gan} to generate hundreds of thousands of synthetic \gls*{lp} images with different transformations and balanced character classes in order to improve the generalization ability of deep models.
Finally, we would like to carry out more experiments to quantify the influence of each dataset, especially \dataset, on the performance of the models under the leave-one-dataset-out~protocol.
\section*{\uppercase{Acknowledgments}}

This work was supported in part by the Coordination for the Improvement of Higher Education Personnel~(CAPES) (Social Demand Program), and in part by the National Council for Scientific and Technological Development~(CNPq) (Grant~308879/2020-1).
The Quadro RTX $8000$ GPU used for this research was donated by the NVIDIA Corporation.
We also thank the \acrfull*{rodosol} concessionaire, particularly the \gls*{it} manager Marciano Calvi Ferri, for providing the images for the creation of the \dataset~dataset.

\balance

\bibliographystyle{apalike}
{\footnotesize
\bibliography{bibtex}}

\end{document}